\documentclass[10pt]{article}

\usepackage[usenames,dvipsnames]{xcolor}

\usepackage{geometry}

\newcommand{\theAuthor}         {-noauthor-}
\newcommand{\theTitle}          {-notitle-}
\newcommand{\theSubtitle}       {-nosubtitle-}

\newcommand{\setAuthor}[1]      {\renewcommand{\theAuthor}{#1}      \author{#1}}
\newcommand{\setTitle}[1]       {\renewcommand{\theTitle}{#1}       \title{#1}}
\newcommand{\setSubtitle}[1]    {\renewcommand{\theSubtitle}{#1}    \subtitle{#1}}

\newcommand{\theAbstractEnglish}{}
\newcommand{\abstractEnglish}[1]{\renewcommand{\theAbstractEnglish}{#1}}

\newcommand{\theAbstractGerman}{}
\newcommand{\abstractGerman}[1]{\renewcommand{\theAbstractGerman}{#1}}

\usepackage{titling}

\usepackage[subfigure]{tocloft}
\usepackage{ccaption}
\usepackage{multicol}

\usepackage{subfigure}

\tocloftpagestyle{fancy}

\usepackage{fancyhdr}
\fancypagestyle{plain}{\pagestyle{fancy}}

\fancypagestyle{plain}{%
\fancyhf{} 

}

\newcommand{\DateYear}{2000}
\newcommand{\DateMonth}{12}

\newcommand{\NumberTechReport}{123}

\newcommand{\setDateYear}[1]            {\renewcommand{\DateYear}{#1}}
\newcommand{\setDateMonth}[1]           {\renewcommand{\DateMonth}{#1}}

\newcommand{\setNumberTechReport}[1]    {\renewcommand{\NumberTechReport}{#1}}

\usepackage[owncaptions]{vhistory}

\newenvironment{versionhistorybottom}
{

\begin{figure}[b!]
\begin{center} 
\footnotesize
\begin{minipage}{0.75\linewidth}
\begin{versionhistory}
}
{
\end{versionhistory}  
\end{minipage}
\end{center}
\end{figure}
}

\makeatletter
\def\s@btitle{\relax}
\def\subtitle#1{\gdef\s@btitle{#1}}
\def\@maketitle{%
  \newpage
  \null
  \vskip 2em%
  \begin{center}%
  \let \footnote \thanks
    {\LARGE \@title \par}%
                \if\s@btitle\relax
                \else\typeout{[subtitle]}%
                        \vskip .5pc
                        \begin{large}%
                                \textsl{\s@btitle}%
                                \par
                        \end{large}%
                \fi
    \vskip 1.5em%
    {\large
      \lineskip .5em%
      \begin{tabular}[t]{c}%
        \@author
      \end{tabular}\par}%
    \vskip 1em%
    {\large \@date}%
  \end{center}%
  \par
  \vskip 1.5em}
\makeatother

\usepackage[citecolor = red]{hyperref}
\usepackage{float}

\usepackage{booktabs}

\usepackage{etex}
\usepackage{tikz}
\usetikzlibrary{matrix,arrows}

\usepackage{amssymb, amsmath}
\usepackage{mathrsfs}
\DeclareMathAlphabet{\mathpzc}{OT1}{pzc}{m}{it}
\usepackage[amsmath, standard]{ntheorem}

\usepackage{blindtext}

\usepackage{lmodern}
\usepackage[utf8]{inputenc}

\usepackage{accents}
\usepackage{fge}
\usepackage{nicefrac}
\usepackage{lmodern}
\usepackage{pifont}

\usepackage{subfigure}
\usepackage{multirow}
\usepackage{appendix}

\usepackage{eso-pic}
\AddToShipoutPictureBG{%
\begin{tikzpicture}[remember picture, overlay]
\node[opacity=1.0, inner sep=0pt]
    at(current page.center){\includegraphics[width=\paperwidth,height=\paperheight]{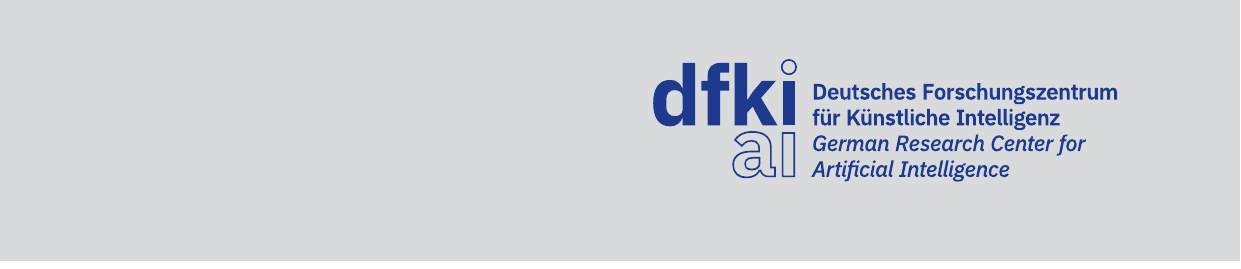}};
\end{tikzpicture}%
}

\abstractEnglish{%
Recent success of machine learning in many domains has been overwhelming,
which often leads to false expectations regarding the capabilities of
behavior learning in robotics.
In this survey, we analyze the current state of machine learning for robotic behaviors.
We will give a broad overview of behaviors that have been learned and used on
real robots. Our focus is on kinematically or sensorially complex robots.
That includes humanoid robots or parts of humanoid robots, for example,
legged robots or robotic arms.
We will classify presented behaviors according to various categories and we
will draw conclusions about what can be learned and what should be learned.
Furthermore, we will give an outlook on problems that are challenging today
but might be solved by machine learning in the future and argue that classical
robotics and other approaches from artificial intelligence should be
integrated more with machine learning to form completely autonomous systems.
}
\abstractGerman{%
Die jüngsten Erfolge des maschinellen Lernens in vielen Bereichen sind überwältigend, was oft zu falschen Erwartungen hinsichtlich der Fähigkeiten des Lernens von Verhalten in der Robotik führt.
In diesem Überblick analysieren wir den aktuellen Stand des maschinellen Lernens für Roboterverhalten.
Wir geben einen breiten Überblick über Verhaltens, die gelernt wurden und auf
realen Robotern genutzt wurden. Unser Schwerpunkt liegt auf kinematisch oder sensorisch komplexen Robotern.
Dazu gehören humanoide Roboter oder Teile von humanoiden Robotern, zum Beispiel Laufroboter oder Roboterarme.
Wir werden die vorgestellten Verhaltensweisen in verschiedene Kategorien einteilen und schlussfolgern was gelernt werden kann und was gelernt werden sollte.
Außerdem geben wir einen Ausblick auf Probleme, die heute noch eine Herausforderung darstellen und die in Zukunft durch maschinelles Lernen gelöst werden könnten und argumentieren, dass die klassische Robotik und andere Ansätze der künstlichen Intelligenz stärker mit maschinellem Lernen integriert werden sollten, um vollständig autonome Systeme zu bilden.
}

\setTitle{A Survey of Behavior Learning Applications in Robotics}
\setSubtitle{State of the Art and Perspectives}
\setAuthor{Alexander Fabisch, Christoph Petzoldt, Marc Otto, Frank Kirchner}

\setDateYear{2024}
\setDateMonth{08}

\setNumberTechReport{RR-24-01}

 \usepackage[ngerman,english]{babel}
\usepackage{pgf}
\usepackage{tikz}

\usepackage{graphicx}
\usepackage{moreverb}
\usepackage{natbib}

\newcommand{\cmark}{\ding{51}}%
\newcommand{\xmark}{\ding{55}}%

\newcommand\BibTeX{{\rmfamily B\kern-.05em \textsc{i\kern-.025em b}\kern-.08em
		T\kern-.1667em\lower.7ex\hbox{E}\kern-.125emX}}

\definecolor{newcolor}{rgb}{0.0, 0.5, 0.0}
\definecolor{newcolor2}{rgb}{0.5, 0.5, 0.0}

\usepackage{todonotes}

\usepackage[strict]{changepage}  
\usepackage{german}

\hypersetup{%
    draft             = false,                                  
    pdftitle          = {\theTitle},                            
    pdfsubject        = {},                                     
    pdfauthor         = {\theAuthor},                           
    pdfkeywords       = {no keywords},                          
    pdfproducer       = {LaTeX with hyperref and thsumbpdf},    %
    colorlinks        = true,                                   
    linkcolor         = darkgray,                               
}

\usepackage{framed}
\begin{document}

%
%
{
\pagenumbering{roman}





\thispagestyle{empty}

\definecolor{dfki_front_page_gray}      {rgb}   {0.35, 0.35, 0.35}

\pgfputat{\pgfxy(-1.75, -0.675)}{\pgfbox[left,base]{ \scalebox{1.4}{\huge \textbf{Research Report}\;\,\textbf{\textcolor{white}{\NumberTechReport}} }}}

{
\color{dfki_front_page_gray}

\pgfputat{\pgfxy(-1.75, -3.5)}{\pgfbox[left,base]{\huge  \textbf{\thetitle} }}
\pgfputat{\pgfxy(-1.75, -4.5)}{\pgfbox[left,base]{ \Large \textbf{\theSubtitle}}}

\pgfputat{\pgfxy(-1.75, -6.5)}{\pgfbox[left,base]{ \large \theauthor, \DateMonth /\DateYear } }

\pgfputat{\pgfxy(-1.75, -22.5)}{\pgfbox[left,base]{
Research Report \NumberTechReport\  \medskip
}}

\pgfputat{\pgfxy(-1.75, -23.0)}{\pgfbox[left,base]{
\textbf{German Research Center for Artificial Intelligence (DFKI) GmbH, Robotics Innovation Center}
}}

} 

\clearpage

{
\thispagestyle{empty}
\setlength{\parindent}{0pt}
\linespread{1.25}
\small

\bigskip

\vfill

\vspace*{3cm}

\begin{minipage}{0.73\linewidth}


\bigskip

© German Research Center for Artificial Intelligence (DFKI) GmbH, 2024

\bigskip
\bigskip

This work may not be copied or reproduced in whole or in part for any commercial
purpose. Permission to copy in whole or in part without payment of
fee is granted for nonprofit educational and research purposes provided
that all such whole or partial copies include the following: a notice that
such copying is by permission of the German Research Center for Artificial
Intelligence (DFKI) GmbH, Kaiserslautern, Federal Republic of Germany; an
acknowledgment of the authors and individual contributors to the work;
all applicable portions of this copyright notice. Copying, reproducing, or republishing
for any other purpose shall require a license with payment of fee
to German Research Center for Artificial Intelligence (DFKI) GmbH.

\bigskip
\bigskip

Issue \NumberTechReport\ (\DateYear)  \\

\end{minipage}

\indent
\clearpage



\begin{titlepage}





\begin{centering}

\vspace*{3cm}

\huge \thetitle

\vspace*{0.5cm}

\Large \theSubtitle

\vspace*{1.5cm}

\Large \theauthor

\vspace*{10cm}

\Large \DateMonth /\DateYear

\normalsize

\vspace*{2cm}

Research Report \NumberTechReport\ des

Deutschen Forschungszentrums für Künstliche Intelligenz (DFKI)

\end{centering}

\end{titlepage}

\clearpage


{
\newpage 
\thispagestyle{empty}
\phantom{some content}
\newpage 
}

\thispagestyle{empty}

\vspace*{\fill}

\section*{\centering Abstract}
\addcontentsline{toc}{section}{Abstract}
\theAbstractEnglish


\vspace*{\fill}


{
\newpage 
\thispagestyle{empty}
\phantom{some content}
\newpage 
}

\tableofcontents

\clearpage

\pagestyle{fancy}

%
%

%
%

\begin{versionhistorybottom}
\vhEntry{0.01}{2018-10-18}{Alexander Fabisch, Christoph Petzoldt, Marc Otto, Frank Kirchner}{Submission to the International Journal of Robotics Research.}
\vhEntry{1.0}{2019-06-05}{Alexander Fabisch, Christoph Petzoldt, Marc Otto, Frank Kirchner}{Preprint published at arXiv.}
\vhEntry{2.0}{2024-05-15}{Alexander Fabisch, Marc Otto}{DFKI research report.}
\end{versionhistorybottom}

\section*{Preface}

This survey was originally submitted to the International Journal of Robotics Research. It was not rejected by the reviewers. However, during the review process the associate editor who was handling it stepped down and he was only able to obtain one review. We received this information almost three years after submission. Since this is a survey paper it was not covering the latest results anymore at that time. However, we did not find the time to integrate new papers in a density with which we covered the research up to submission to resubmit it. Since the survey was published as a preprint on arXiv for several years, it was also cited several times and we believe the conclusions still stand today. Hence, we finally decided to publish it officially as a research report at DFKI.

In addition to the generally positive review that we got from the journal, this survey was also thoroughly reviewed by our colleagues Thomas M. Roehr and José de Gea Fernández in the internal review process of the DFKI Robotics Innovation Center.

We want to provide the original survey as of the state of October 2018. However, we would like to make one additional point considering recent developments in machine learning and robot learning in particular: the current trend in machine learning and robot learning favors expensive, large models pretrained on exorbitant datasets \citep{Bommasani2021,OpenXEmbodiment2023,Ajay2023Is,Huang2023Skill,Khazatsky2024Droid}, in which basic physical plausibility is not guaranteed \citep{Liu2024Sora}. We believe that this trend is concerning and we argue that future research should focus more on sample-efficiency, low resources, low computational cost, and green AI \citep{Schwartz2020Green}.

}

\clearpage
%
%
{
\pagenumbering{arabic}
%
%

\pagebreak

\section{Introduction}


Machine learning and particularly deep learning \citep{LeCun2015} made
groundbreaking success possible in many domains,
such as computer vision \citep{Krizhevsky2012},
speech recognition \citep{Hinton2012}, playing video games \citep{Mnih2015},
and playing Go \citep{Silver2016}.
It is unquestionable that learning from data, learning from experience and
observations are keys to really adaptive and intelligent agents -- virtual
or physical.
However, people are often susceptible to the fallacy that the state of the
art in robotic control today heavily relies on machine learning.
This is often not the case.
An example for this is given by \citet{Irpan2018}:
at the time of writing this paper, the humanoid robot Atlas from Boston
Dynamics is one of the most impressive works in robot control.
It is able to walk and run on irregular terrain, jump precisely with one
or two legs, and even do a back flip \citep{BostonDynamics2018}.
\citet{Irpan2018} reports that people often assume that Atlas uses
reinforcement learning.
Publications from Boston Dynamics are sparse, but they do not include
explanations of machine learning algorithms for control
\citep{Raibert2008,Nelson2012}.
\citet{Kuindersma2016} present their work with the robot Atlas, which
includes state estimation and optimization methods for locomotion behavior.
Robotic applications have demanding requirements on processing power,
real-time computation, sample-efficiency, and safety, which often makes the
application of state-of-the-art machine learning for robot
behavior learning difficult.
Results in the area of machine learning are impressive but they
can lead to false expectations.
This led us to the question: what can and what should be learned?


Recent surveys of the field mostly focus on algorithmic aspects of machine
learning \citep{Billard2008,Argall2009,Kober2013,Kormushev2013,Tai2016,Arulkumaran2017,Osa2018}.
In this survey, we take a broader perspective to analyze the state of the
art in learning robotic behavior and do explicitly not focus on algorithms
but on (mostly) real world applications.
We explicitly focus on applications
with real robots, because it is much more demanding to integrate and learn
behaviors in a complex robotic system operating in the real world.
We give a very broad overview of considered behavior learning problems on
real robotic systems.
We categorize problems and solutions, analyze problem characteristics,
and point out where and why machine learning is useful.

This article is structured as follows.
We first present a detailed summary of selected highlights that advanced the
state of the art in robotic behavior learning.
We proceed with definitions of behavior and related terms.
We present categories to distinguish and classify behaviors
before we present a broad overview of the state of the art in
robotic behavior learning problems.
We conclude with a discussion of our findings and an outlook.

\section{Selected Highlights}
\label{s:highlights}

Among all the publications that we discuss here, we selected some
highlights that we found to be relevant extensions of the repertoire of
robotic behavior learning problems that can be solved.
We briefly summarize these behavior learning problems and their solutions
individually before we enter the discussion of the whole field from a broader
perspective.
We find it crucial to understand the algorithmic development and
technical challenges in the field. It also gives a good impression of
the current state of the art.
Later in this article, we make a distinction whether the perception or the
action part of these behaviors have been learned (see Figure
\ref{fig:perception_action}).


An early work that combines behavior learning and robotics has been
published by \citet{Kirchner1997}.
A goal-directed walking behavior for the six-legged walking machine
SIR ARTHUR with 16 degrees of freedom (DOF) and four light sensors
has been learned.
The behavior has been learned on three levels --
(i) bottom: elementary swing and stance movements of individual
legs are learned first, (ii) middle: these elementary actions are then
used and activated in a temporal sequence to perform more complex
behaviors like a forward movement of the whole robot,
and (iii) top: a goal-achieving behavior in a given environment with
external stimuli.
The top-level behavior was able to make use of the light sensors to
find a source of maximum light intensity.
Reinforcement learning, a hierarchical version of Q-learning
\citep{Watkins1989}, has been used to learn the behavior.
On the lowest level, individual reward functions for lifting up the leg,
moving the leg to the ground, stance the leg backward, and swinging the
leg forward have been defined.


\citet{Peters2005} presented an algorithmic milestone in reinforcement
learning for robotic systems.
They specifically used a robot arm with seven degrees of freedom (DOF)
to play tee-ball, a simplified version of baseball, where the ball
is placed on a flexible shaft.
Their solution combines imitation learning through kinesthetic teaching
with dynamical movement primitives (DMPs) and policy search, which
is an approach that has been used in many following works.
In their work, \citet{Peters2005} used natural actor-critic (NAC) for policy search.
The goal was to hit the ball so that it flies as far as possible.
The reward for policy search included a term that penalizes squared accelerations and
rewards the distance.
The distance is obtained from an estimated trajectory computed with
trajectory samples that are measured with a vision system.
An inverse dynamics controller has been used to execute motor commands.
About 400 episodes were required to learn a successful batting behavior.


Ball-in-a-cup is a very challenging game. A ball is attached to a cup
by a string. The player has to catch the ball with the cup by moving
only the cup. Even human players require a significant amount of trials
to solve the problem.
\citet{Kober2008,Kober2009} demonstrate that a successful behavior can be
learned on a SARCOS arm and a Barret WAM.
A similar approach has been used: imitation learning with DMPs from
motion capture or kinesthetic teaching and refinement with a policy
search algorithm, in this case Policy Learning by Weighting Exploration
with the Returns (PoWER). In addition, the policy takes the ball position into
consideration. A perceptual coupling is learned to
mitigate the influence of minor perturbations of the end-effector
that can have significant influence on the ball trajectory.
A successful behavior is learned after 75 episodes.


The problem of flipping a pancake with a pan has been solved by
\citet{Kormushev2010b} with the same methods: a controller that is very
similar to a DMP is initialized from kinesthetic teaching and refined
with PoWER.
The behavior has been learned with a torque-controlled Barrett WAM arm
with 7 DOF.
The artificial pancake has a weight of 26 grams only, which makes its motion less
predictable because it is susceptible to the influence of air flow.
For refinement, a complex reward function has been designed that
takes into account the trajectory of the pancake (flipping and catching),
which is measured with a marker-based motion capture system.
After 50 episodes, the first successful catch was recorded.
A remarkable finding is that the learned behavior includes a useful aspect that
has not directly been encoded in the reward function: it made a compliant
vertical movement for catching the pancake which decreases the chance
of the pancake bouncing off from the surface of the pan.


Table tennis with a Barrett WAM arm has been learned by
\citet{Muelling2011,Muelling2013}. Particularly challenging
is the advanced perception and state estimation problem. In comparison to
previous work, behaviors have to take an estimate of the future
ball trajectory into account when generating movements that determine where, when, and how the robot hits the ball.
A vision system has been used to track the ball with 60 Hz.
The ball position is tracked with an extended Kalman
filter and ball trajectories are predicted with a simplified model that
neglected the spin of the ball. 25 striking movements have been learned from
kinesthetic teaching to form a library of movement primitives.
A modified DMP version that allows to set a final velocity as a meta-parameter
has been used to represent the demonstrations.
Desired position, velocity and orientation of the racket are computed
analytically for an estimated ball trajectory and a given target on
the opponent's court and are given as meta-parameters to the modified
DMP.
In addition, based on these task parameters, a weighted average of known
striking movements is computed by a gating network.
This method is called mixture of movement primitives.
The reward function encourages minimization of the distance between the desired
goal on the opponent's court and the actual point where the
ball hits the table.
In the final experiment, a human played
against the robot, serving balls on an area of 0.8 m $\times$ 0.6 m.
Up to nine balls were returned in a row by the robot. Initially the
robot was able to return 74.4\,\% of the balls and after playing one
hour the robot was able to return 88\,\%.


Learning end-to-end behaviors that take raw camera images to compute
corresponding motor torques (visual servoing) has been demonstrated
impressively by \citet{Levine2016}.
They use the 7 DOF arm of a PR2 robot to learn a variety of isolated manipulation behaviors:
hanging a coat hanger on a clothes rack, inserting a block into a shape sorting
cube, fitting the claw of a toy hammer under a nail, and
screwing a cap on a water bottle.
The final behaviors use a convolutional neural network (CNN) to control the
arm's movements at 20 Hz based on the visual input from a monocular RGB camera
with a resolution of 240x240 pixels.
A sophisticated training process involving several phases has been developed in this work.
The first layer of the convolutional neural network is initialized
from a neural network that has been pretrained on ImageNet \citep{Deng2009}.
In a second pretraining step, the image processing part of the neural network
is initialized by training a pose regression convolutional neural network
that predicts 3D points that define the target objects involved in the task.
Guided policy search is used to train the final policy.
The whole state of the system is observed during this training phase
and a local dynamic model is trained. An optimal control method
that uses the full system state is used
to obtain a ``guiding policy''. This guiding policy is used to
train the neural network policy in a fully supervised setting.
The final neural network policy, however, works directly on images that
represent partial information about the state of the system without
having the knowledge of the full system state that would only be available
during training.
The whole training process for a new behavior requires 3--4 hours.


Grasping has also been learned from raw monocular RGB camera images
with a 7 DOF robot arm by \citet{Levine2017}. In this application,
the behavior is not learned end-to-end, but a neural network
has been learned to predict the success of a motion command for
a given camera image (and the camera image before the behavior is started).
The behavior goes through a sequence of ten waypoints defined by the
Cartesian end-effector position and the rotation of the 2-finger gripper
around the z-axis. A motion command is selected in each step by an
optimization procedure based on the predicted success of the motion command.
The remarkable fact about this work is that a total amount of more than
800,000 plus 900,000 grasps collected in two datasets have been performed to
train the grasp success prediction model and a maximum of 14 robots has been
used in parallel to collect the data. A large variety of objects
has been used to test the learned grasping behavior.


\section{Definition of Behavior}
\label{s:definition}

Before we enter the discussion of robotic behaviors, we clarify several
related terms. These are mostly taken from biology.

We borrow a definition of the term \textbf{behavior} from behavioral
biology. Unfortunately, many behavioral biologists disagree in the definition
of behavior \citep{Levitis2009}. Hence, we will select one and this is the one
proposed by \citet{Levitis2009}:
``behaviour is the internally coordinated responses (actions or inactions)
of whole living organisms (individuals or groups) to internal and/or external
stimuli\,..''.
Note that we excluded a part of the original definition as it only
applies to biological systems.
For our purposes we extend this definition to artificial systems like robots.
Furthermore, \citet{Levitis2009} point out ``Information processing may be a
necessary substrate for behaviour, but we do not consider it a behaviour by
itself.'' This is an important statement because it excludes perception,
state estimation, and building world models from the definition of behavior
while it may be part of a behavior.

There are other terms related to behavior and behavior learning that we
use in the discussion.
\citet[page 46]{Shadmehr2005} state
``Once the CNS [central nervous system] selects the targets (or goals) of
reach ... it must eventually compute a motor plan and generate the coordinated
forces needed to achieve the goal, even if this computation evolves during the
movement. The ability to achieve such goals typically requires a motor skill.''
Hence, we can distinguish the more general concept of a motor skill
and an explicit and specific motor plan.
The term \textbf{skill} is widely used.
We define skill as a learned ability of an organism or artificial system.
A skill is not the behavior but a behavioral template that can be adapted
to a behavior for certain situations that are similar to those in which it
was learned.
A set of skills constitutes a skill library or motor repertoire.
A \textbf{motor plan} is a sequence of actions to be taken in order to achieve
a given goal.
Another term that is often used in the context of robot skill learning
is \textbf{movement primitive}. Movement primitives are
``fundamental units of motor behavior'', more precisely,
``indivisible elements of motor behavior that generate a regulated and stable
mechanical response'' \citep{Giszter1993}.
More specifically, a movement primitive can represent a learned skill and
a motor plan is a skill adapted to a specific situation.


\section{Classification of Behaviors}
\label{s:classification}

Now that we have defined behavior and related terms,
we will introduce categories to distinguish and classify
behaviors and behavior learning problems.
Note that some behaviors cannot clearly be categorized or
some categories do not even apply to all behaviors.
In contrast to \citet{Schaal2010}, we focus completely
on classifying the problem and corresponding behavior, not on
the method that is used to solve the problem or generate
the behavior, and we use more refined categories to
characterize these behaviors.

\paragraph{Domain:}
Behaviors are often useful only in specific domains. Sometimes similar
but different behaviors are used in different domains. Examples for domains
are manufacturing, healthcare, logistics, household, or games.
We will explicitly exclude military applications.
Here, we will follow a bottom-up approach to identify relevant domains that
include a significant amount of learned behaviors.


\paragraph{Hierarchy of behaviors:}
Behaviors can have different timescales and levels of abstraction regarding
goals. For example, keeping a household clean is more abstract and
time-consuming than picking up a particular cup.
Furthermore, behaviors can consist of sub-behaviors, as shown in
Figure~\ref{fig:behavior_hierarchy}.
A resource management behavior can achieve the goal of maintaining a storage
filled by keeping track of the stored amount (stocktaking) and collecting
resources (foraging) when necessary.
As goals become more concrete and faster to achieve, their priority generally
increases: in the example, keeping balance or avoiding an obstacle are often
obligatory leading to compromises in the achievement of higher level goals.
%
Sub-behaviors may be executed in parallel or in a sequence and generally,
the type of their combination (output weighting, suppression, sequence) is
learnable.

Organizing behaviors hierarchically has been demonstrated to be of
practical relevance to organize hand-coded behaviors for the complex
domain of robot soccer. The behavior specification languages
XABSL \citep{Loetzsch2006} and CABSL \citep{Roefer2018} are common
among robot soccer teams.
A hierarchical behavior structure is also useful to divide the learning
procedure, as demonstrated by \citet{Kirchner1997}.
Hierarchical behavior organization dates back at least to the field of
behavior based robotics \cite{Arkin1998}, manifested, for example,
in the subsumption architecture of \citet{Brooks1986}.

\paragraph{Perception and action:}
Behaviors often involve perception and action (see Figure
\ref{fig:perception_action}).
Some behaviors can be executed open-loop. They do not incorporate any
sensory feedback after they have been started. Pure perception
on the other hand does not match our definition of behavior. However,
often a coupling between perception and action is required. Sometimes
both components are learned, sometimes only the action is learned and
sometimes there is a stronger focus on learning the perception part of
the behavior. We will indicate which part of the behaviors are learned
with this classification.

\begin{figure}[!tb]
\centering
\includegraphics[width=0.48\textwidth]{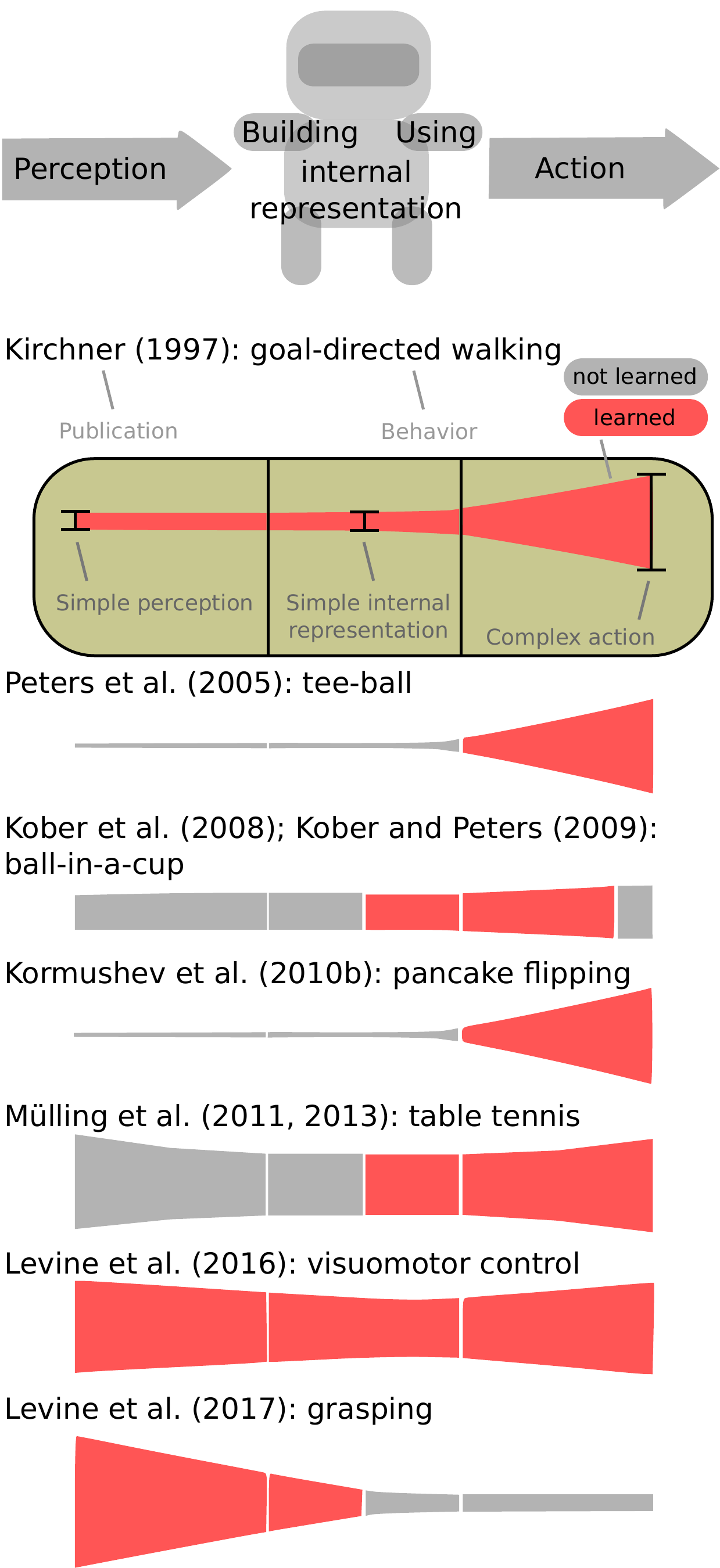}
\caption{
\textbf{Perception and action.} The red background indicates which parts of the
behavior are learned. Sometimes both, perception and action, are learned and sometimes only
some aspects are learned. The height of each bar indicates complexity of
the corresponding part.\label{fig:perception_action}}
\end{figure}

\paragraph{Deliberative vs. reactive behaviors:}
\citet{Arkin1998} distinguishes between deliberative and reactive robot
control. This can be transferred directly to robotic behavior.
Deliberative control often relies on a symbolic world model. Perception
is not directly coupled to action, it is used to populate and update the
world model. Actions are derived from the world model. Deliberative control
is usually responding slowly with a variable latency and can be regarded as
high-level intelligence. We define deliberative behaviors as behaviors that
only have an indirect coupling between sensors and actuators through a form
of world model. Behaviors that are learned completely are usually not
deliberative. Only parts of deliberative behaviors are learned.
Reactive control does not rely on a world model because it couples perception
and action directly. It usually responds in real-time, relies on simple
computation, and is a form of low-level intelligence. Reactive control
architectures often combine multiple reactive behaviors. An interesting
property of these architectures is that often unforeseen high-level behavior
emerges from the interplay between robot and environment. Reflexive behavior
is purely reactive behavior with tight sensor-actuator coupling.
Deliberative and reactive behaviors are often closed-loop behaviors.
Behaviors without coupling between perception and action also exist.
These are open-loop behaviors.
Sometimes open loop behaviors are triggered with a hard-coded rule based
on sensor data. 
Note that sensor data used during the training phase is irrelevant for
this classification, only sensor data during execution of the behavior
is relevant.

\paragraph{Discrete vs. rhythmic behavior:}
\citet{Schaal2004} distinguish between two forms of movements:
discrete and rhythmic movements. Discrete movements are point-to-point movements
with a defined start and end point. Rhythmic movements are periodic without
a start or end point or could be regarded as a sequence of similar discrete
movements. Some behaviors might be rhythmic on one scale and discrete on
another scale. This distinction has often been used for robotic behaviors.
Hence, we adopt it for our survey. \citet{Schaal2004} show that discrete
movements often involve higher cortical planning areas in humans and propose
separate neurophysiological and theoretical treatment.

\paragraph{Static vs. dynamic behavior:}
We introduce a classification of behaviors that distinguishes between
dynamic behavior and static behavior. Momentum is very important in
dynamic behaviors because it will either be transferred to
the environment or it is required because the robot or the
environment is not stable enough to maintain its state without momentum.
Static behaviors can be interrupted at any time and then continued
without affecting the outcome of the behavior. In practice,
some behaviors also lie in between, because momentum is not important
but interrupting the behavior might alter the result insignificantly.
Some problems would usually be solved by a human with dynamic behaviors
but when the behavior is executed slow enough, it loses its dynamic
properties. This is often the case when robots solve these kinds of problems.
We call these kind of behaviors quasi-static.
This categorization is inspired, for example, by research in walking robots:
a static walk can be stopped at any time and the robot will stay
indefinitely at the same position \citep{Benbrahim1997}.
A similar categorization into dynamic and static movement techniques
is made in rock climbing \citep{WikiClimbing2018}.
A complementary definition for manipulation is provided by \citet{Mason1993}:
static manipulation is defined as an operation ``that can be
analyzed using only kinematics and static forces'', quasi-static
manipulation can be analyzed ``using only kinematics, static forces,
and quasi-static forces (such as frictional forces at sliding contacts)'',
and dynamic manipulation can be analyzed ``using kinematics, static and
quasi-static forces, and forces of acceleration''.

\paragraph{Active vs. passive:}
Some behaviors are executed with the intention to actively change
the state of the robot or the world. Others are only passive
and often have the goal of maintaining a state like homeostasis,
that is, a state of steady internal conditions.
Change of the environment is a side effect. We borrow this idea
from the behavior architecture of \citet{Rauch2012} but it can
be applied to any level of behavior.

\paragraph{Locomotion vs. manipulation:}
Many implemented behaviors of existing robotic systems can be categorized
as locomotion or manipulation.
Locomotion includes all behaviors that move the robot and, thus, change the
state of the robot in the world. Change of the environment is a side effect.
Manipulation behaviors change the state of the environment. Changing the state
of the robot is a side effect. Manipulation is typically characterized as
mechanical work that modifies the arrangement of objects in the world.

\paragraph{System requirements:}
Behaviors have different requirements on the hardware design of the robot.
Many locomotion behaviors require legs,
manipulation behaviors require grippers, hands,
and / or arms.
Navigation and exploration behaviors often only require wheels.
Some behaviors rely on particular sensors, for example, cameras, force-torque
sensors, or distance sensors. We will mention the most important requirements
in the description of the behaviors if they are not obvious. An example of
an obvious requirement is that a walking robot needs
something similar to legs.

\paragraph{Noise and uncertainty:}
Behavior learning applications are significantly more difficult if there is
noise in state transitions or state perception. Sometimes the state is
not fully observable and, hence, there is uncertainty in perception.
Sometimes the state transition is not fully determined by the actions that
the robot can execute because the environment itself is dynamic. This is
another reason for uncertainty.


\section{Robotic Behavior Learning Problems}
\label{s:problems}

Robotic behaviors can be learned with many different approaches. Two relevant
branches are reinforcement learning and supervised learning.
Recent surveys on reinforcement learning in robotics have been published
by \citet{Kober2013,Kormushev2013}.
Deep reinforcement learning is a new field that makes use of the results from deep learning.
Although there are only a few applications of deep reinforcement learning
in robotics, results of these methods are interesting for behaviors that involve difficult
perception problems. A recent survey of deep reinforcement learning has been
published by \citet{Arulkumaran2017} and a survey of deep learning for
robotic perception and control by \citet{Tai2016}.
Supervised learning can be used to learn the perception
part of a behavior, the action part, or both. If actions are learned supervised,
this is called imitation learning or programming by demonstration. Surveys have
been written by \citet{Billard2008,Argall2009,Osa2018}.
We do not discuss algorithms in this section.
Please refer to these surveys or to other papers that we cite in this
section to learn more about specific algorithms that can be used to learn
behaviors.
We neither discuss the reported performance of the solutions
from the presented works.


We will focus on kinematically or sensorially complex robots.
That includes humanoid robots or parts of humanoid robots like
legged robots or robotic arms.
We only consider applications for unmanned aerial vehicles,
autonomous underwater vehicles, or wheeled robots
if the learned behaviors are relevant for humanoid robots.
That excludes some early works that apply machine learning
to robotic control, for example, \citet{Mahadevan1992} learn a behavior
to find and push a box with a wheeled robot, but also more recent work
with deep reinforcement learning on robotic systems.
We also do not discuss behaviors that have only been demonstrated in
simulation because of the reality gap \citep{Jakobi1995}.



In this section, we try to capture the large variety of robotic behavior
learning problems according to the presented definition of behavior.
We group problems according to the categories introduced in the previous
section and point out similarities and differences between and difficulties
of these problems.

A histogram that shows the distribution of the analyzed papers by publication
dates is displayed in Figure \ref{fig:years}.
Although we do not claim to have included definitely every relevant work,
it shows that the number of applications of behavior learning to robotic
systems has been growing fast in the last 10 years.
Figure \ref{fig:mindmap} shows how individual behaviors can be grouped by
their domain of application. Some behaviors can of course be applied in
several domains. These are elementary behaviors. Examples are walking and
grasping.
Table \ref{tab:complete} summarizes the behavior learning problems,
corresponding publications, and their categorization.
The remainder of this section is separated in manipulation behaviors,
locomotion behaviors, and behaviors that do not fit any of these categories.

\begin{figure}
\centering
\includegraphics[width=0.48\textwidth]{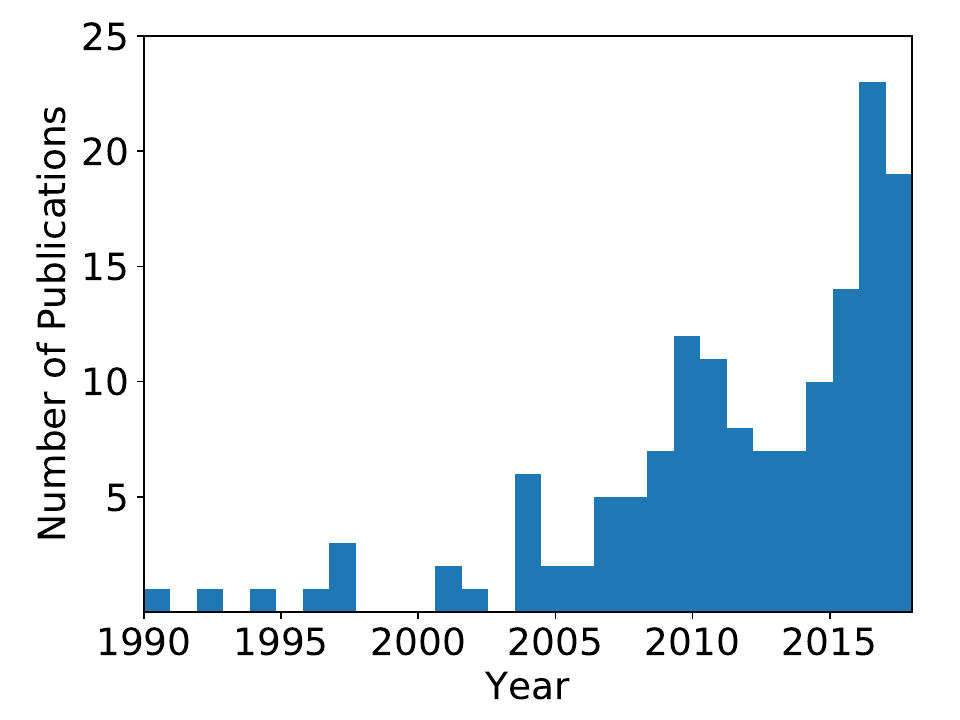}
\caption{\textbf{Histogram of publication years of the considered works.}\label{fig:years}}
\end{figure}

\begin{figure*}
\centering
\includegraphics[width=0.92\textwidth]{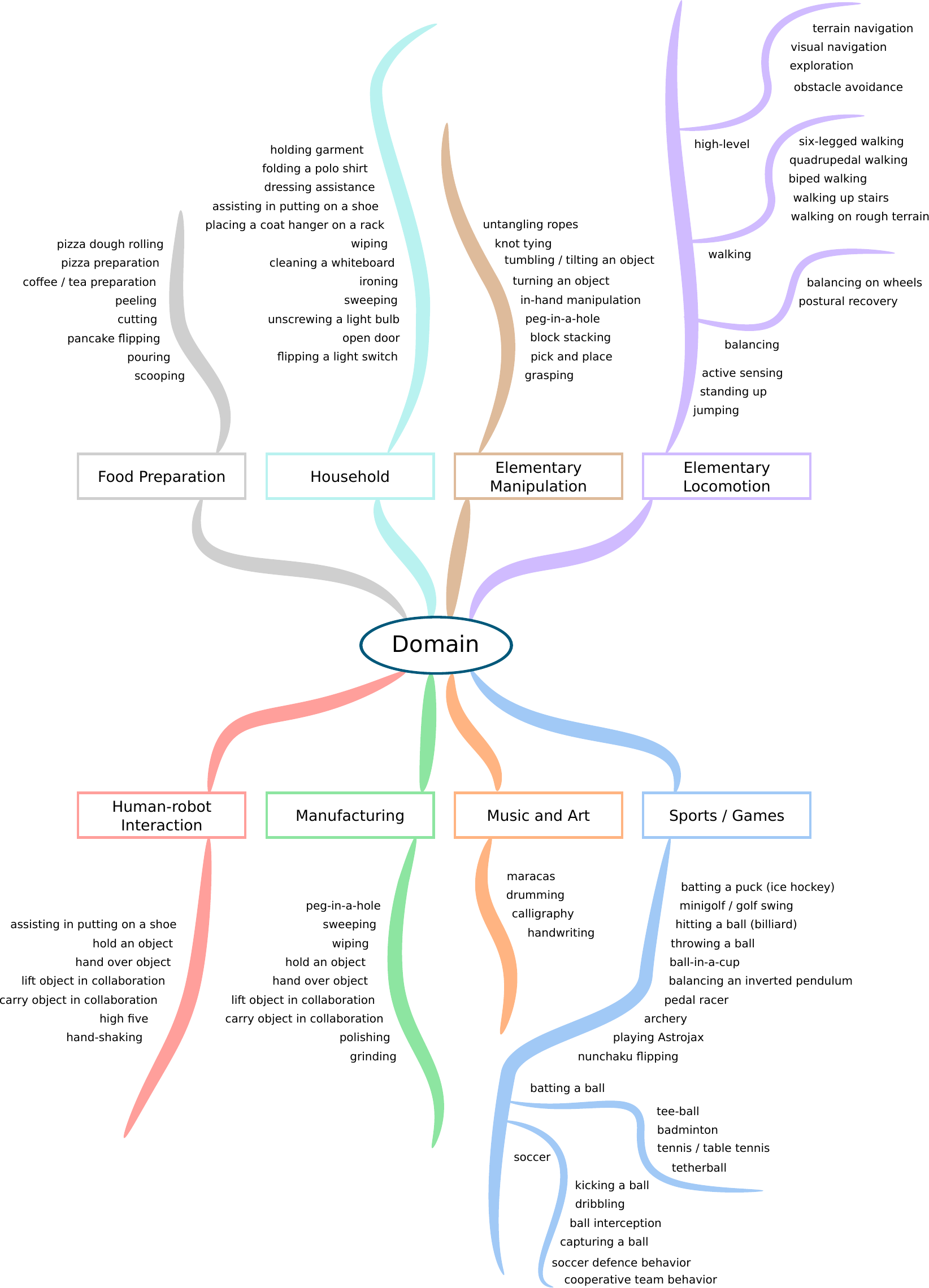}
\caption{
\textbf{Mindmap of behavior learning applications.}
Applications are ordered by domain.
Some behaviors are assigned to multiple domains and most of the elementary
behaviors could also belong to multiple domains.\label{fig:mindmap}}
\end{figure*}

\subsection{Manipulation Behaviors}

Figure \ref{fig:manipulation_overview} shows the categorization of
manipulation behaviors that we used to structure this section.
Manipulation behaviors change the state of the robot's environment,
hence, we categorized behaviors by the softness of the manipulated
object and the dynamics of the behavior.
This is similar to how \citet{Sanchez2018} structured their survey
about manipulation and sensing of deformable objects.
We found this categorization to be useful to organize publications
that we present here. It might, however, not be easily applicable in
other cases. For example, in case of a robot that moves a catheter
\citep{Tibebu2014}, we would have to answer the question if the
catheter is the manipulated object or part of the robot. If the
catheter is part of the robot, what would be the manipulated object?


\begin{figure*}[tb]
\centering
\includegraphics[width=\textwidth]{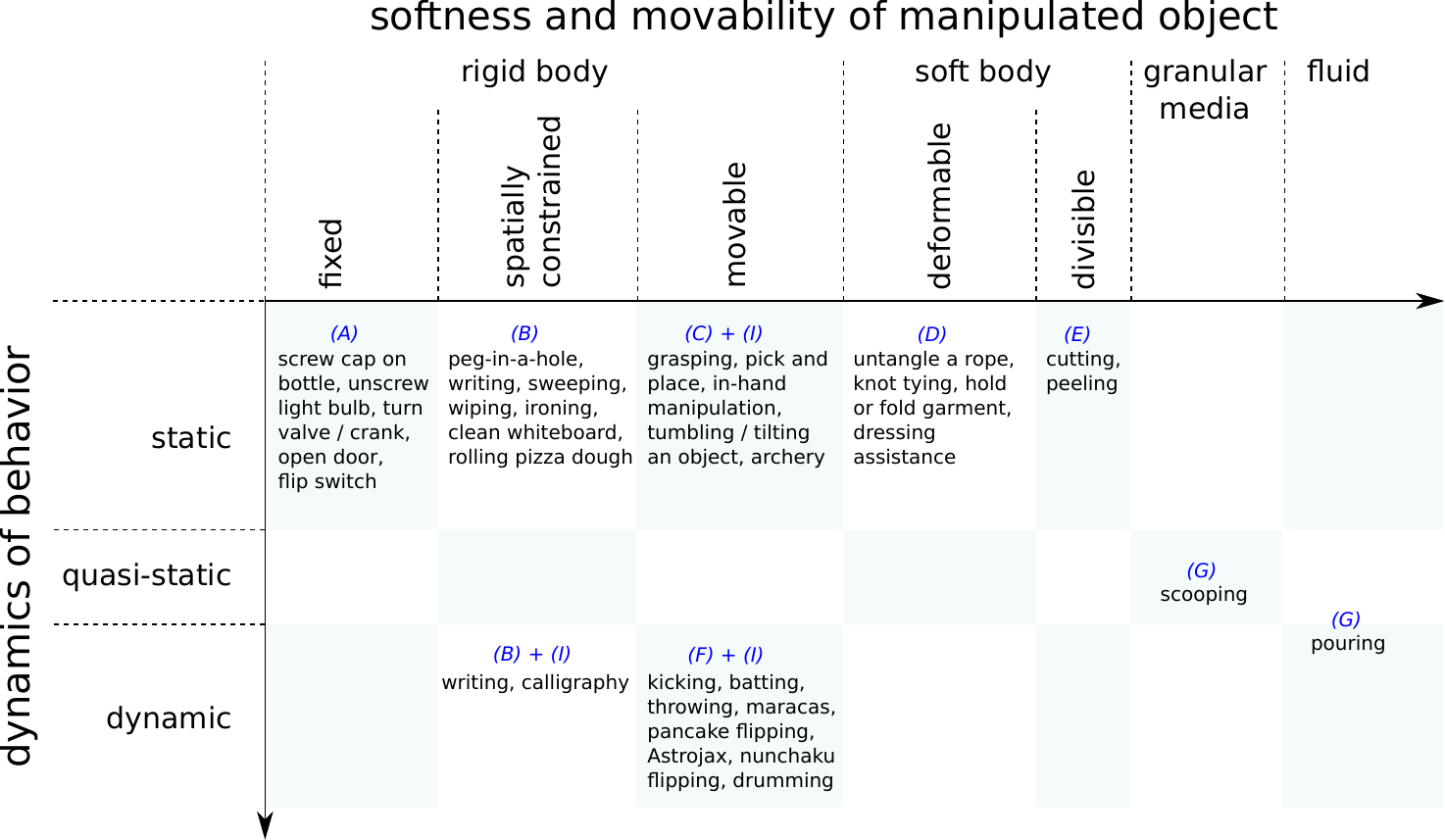}
\caption{
\textbf{Categorization of manipulation behaviors.}
Manipulation behaviors are categorized in two dimensions:
softness and movability of the manipulated object and dynamics of the
behavior. Blue letters indicate the corresponding
subsections.\label{fig:manipulation_overview}
}
\end{figure*}

\subsubsection{Fixed Objects (A)}

\paragraph{Flipping a light switch:}
\citet{Buchli2011} investigate the task of flipping a light switch.
The switch essentially is a via-point that has to be passed through
very precisely in this kind of task.
In addition to high accuracy, the flipping process itself requires the exertion
of forces.
In their work, the robot learns to be compliant when it can be and be stiff
only when the task requires either high precision or exertion of forces.
The problem could be extended to the recognition of the switch, which is not
done here.

\paragraph{Open door:}
In contrast to flipping a switch, opening a door does not require precise
trajectories. Additionally, more than just a via-point problem has to be
solved: opening a door involves grasping the handle, closing the kinematic
chain between gripper and the handle and finally moving the handle.
The movements of the robot after grasping are
restricted by the structure of the handle. Opening a door requires significant
force exertion from the robot to the environment.
\citet{Nemec2017} ignore the problem of grasping and only consider the problem
of learning the unconstraint DOFs while the kinematic chain from the robot
to the door is closed. \citet{Chebotar2017b,Gu2016} consider
the problem of learning this behavior end-to-end from camera images to
motor torques. \citet{Nemec2017,Englert2018} ignore the perception part of
the problem and assume known relative positions.
\citet{Kalakrishnan2011,Kormushev2011} use force sensors. The door
considered by \citet{Kormushev2011} does not have a handle but a horizontal
bar that has to be pushed with a larger force than a standard door handle.
It is also the only work in which the door has been pushed and not pulled.
\citet{Nemec2017,Englert2018} consider not only horizontal but also vertical
handles.


\paragraph{Turning objects:}
Several manipulation problems involve turning fixed objects, for example,
turning a valve \citep{Carrera2012}, or a crank \citep{Petric2014}, or
screwing a cap on a (pill or water) bottle \citep{Levine2016}.
The challenge is to reach a via-point and then hold and move an object on a
circular path.
These behaviors can be realized as rhythmic movements \citep{Petric2014}
or discrete movements \citep{Carrera2012,Levine2016}. They can be discrete
when the object has to be turned only by a small angle (for example, 90
degrees, \citet{Carrera2012}) or when the robot can spin its wrist
\citep{Levine2016}. Some works focus more on robustly reaching the
target object \citep{Carrera2012,Levine2016} and others on robustly turning
the object itself \citep{Petric2014}. \citet{Carrera2012} exclude perception
from learning, \citet{Levine2016} learn perception and action, and
\citet{Petric2014} follow previously learned torque profiles.




\subsubsection{Spatially Constrained Behavior (B)}

\paragraph{Peg-in-a-hole:}
Inserting a peg in a hole is one of the most basic manipulation skills that
we discuss in this article.
It is the most frequent assembly operation \citep{Gullapalli1994}.
The behavior can benefit from both visual \citep{Levine2016} and
force sensors \citep{Gullapalli1994,Ellekilde2012,Kramberger2016}, but it can
also be done without any sensors \citep{Chebotar2017}.
While the most obvious application of this
skill is found in assembly tasks \citep{Gullapalli1994,Ellekilde2012,Kramberger2016,Levine2016},
it can also be used to, for example, plug in a power plug \citep{Chebotar2017}.
The problem can be solved end-to-end from visual data to motor torques
\citep{Levine2016} or from force measurements to Cartesian positions
\citep{Gullapalli1994} as a purely reactive behavior.
Alternatively, learning can be combined with search heuristics for the
hole based on force measurements \citep{Ellekilde2012,Kramberger2016}.
In the simplest case, the behavior is learned for a fixed relative
transformation between robot and target \citep{Chebotar2017}.

A more advanced assembly operation that involves multiple instances of the
peg-in-a-hole problem has been learned by \citet{Laursen2018} to
connecting a pipe for a heating system. In this task, a passively compliant
gripper holds a tool extension and has to use a tube feeder, nut feeder, and
crimping machine. Only actions were learned and a safety mechanism prevented
the system from serious collisions. Apart from that, the system learns
blindly without any sensors.


\paragraph{Wiping:}
The motion required to solve sweeping, wiping, ironing or whiteboard cleaning
tasks can be either discrete or rhythmic. Further, all these task require
environmental interaction by exerting (specific) forces on external objects.
Learning mostly focuses on finding parameters for the representation of the movement.
\citet{Kormushev2010,Kormushev2011} let a robot learn a discrete
ironing skill from demonstrated trajectories and additional force profiles.
They also evaluated their work on a whiteboard cleaning task
\citep{Kormushev2011b}. A similar task
is surface wiping which is investigated by \citet{Urbanek2004,Gams2014}.
Both works represent the wiping skill as a periodic movement.
In this case, rhythmic motions are
advantageous, as the complete surface can be wiped easily by executing
the motion several times while shifting only the center point. The
work from \citet{Gams2014} also uses force feedback to maintain contact
with the surface.
Besides the aforementioned household tasks, there are also industrial
operations that require constant environmental contact.
From these, grinding and polishing tasks have been investigated by \citet{Nemec2018}.
The goal of these tasks is to keep contact with a specific force exertion
between a polishing/grinding machine and the treated object,
which is manipulated by a robot with a desired orientation.
Therefore, their approach reproduces the relative motion between object
and tool.
The contact point is estimated using measured the forces and torques and
can be changed to optimize a defined criterion, for example, minimize joint
velocities.
Sweeping has been considered by \citet{Alizadeh2014}. The position of
``dust'' is obtained using computer vision and the behavior is adapted
accordingly.
\citet{Pervez2017} train a sweeping behavior end-to-end from visual inputs
to collect trash placed at various positions between a fixed initial and goal
position.


\paragraph{Handwriting:}
The goal of handwriting tasks is to resemble human writing as precise and
smooth as possible.
Complete words have been reproduced and generalized on real robots:
\citet{Manschitz2018} learn to generalize a handwriting skill to unseen
locations of a whiteboard which is defined as the target writing position.
\citet{Berio2016} learn to dynamically draw graffiti tags. In comparison
to the above mentioned behavior, these drawings particularly require fluid and
rapid manipulation of the pen to produce elegant and smooth sequences of
letters. Precision is less important for this behavior.

\subsubsection{Movable Objects (C)}

\paragraph{Grasping:}
Grasping is a good example for a high diversity of similar but different
task formulations.
The problem of grasping is usually tightly coupled with perception, but it
can be separated into perception and movement generation.
Continuous feedback can be used to verify the grip although it can also be
sufficient to perceive the target before the grasp attempt.
Problem formulation for grasping varies in the degree of automation and amount
of other methods used in the process. Sometimes perception is learned and
movement generation is done with other approaches and vice versa.
Some approaches learn full reaching and grasping movements for known object
locations \citep{Grave2010,Kalakrishnan2011,Stulp2011,Amor2012},
others just learn to predict grasp poses \citep{Lenz2015,Johns2016,Pinto2016}.
\citet{Steil2004} only consider the problem of defining hand postures
and \citet{Kroemer2009} the problem of learning hand poses relative to
objects.
A full grasping movement includes a reaching trajectory, positioning the
gripper at the correct position, closing the gripper, and sometimes
objects have to be moved in the right position before the gripper can
be closed.
From the works that are mentioned here,
\citet{Grave2010,Steil2004,Stulp2011} do not learn to use
feedback from sensors, \citet{Kroemer2009} use features obtained from
images, \citet{Kalakrishnan2011} use force measurements,
\citet{Lenz2015} use RGB-D images, \citet{Johns2016,Mahler2017} use depth
images, and \citet{Lampe2013,Pinto2016,Levine2017} use RGB images.
Figure \ref{fig:grasping} illustrates possible inputs and outputs
of a component that generates grasping behavior.
A classification proposed by \citet{Bohg2014} distinguishes between grasping
of known, familiar, and unknown objects. Familiar means that the robot did
not encounter the objects before, but has seen similar objects.
Most of the works that we present here fall into this category.
For grasping, other factors that influence the difficulty of the
problem are the used hand or gripper and the objects that should
be grasped.
Very promising results are shown by \citet{Levine2017,Mahler2017}.
A large variety of different objects can be grasped with a
two-finger gripper just based on images or depth images respectively.
However, there are still many options for improvements.
The gripper can only grasp objects with top-down movements.
In the real world, not all problems can be solved with these
kind of grasps.
The gripper only has two fingers. Hands with more fingers
have better control over grasped objects.
Using force feedback and tactile sensors would certainly
improve grasping in some situations.
In a box full of objects, the approach of \citet{Levine2017} just
picks a random object. In practice, this should be a parameter of
the behavior.
Also, it is not clear where and in which orientation the gripper
holds the object.
This does not seem to be a problem because most works just consider
the grasping phase but not what happens afterwards.
In a real application, most probably the object will have to be
placed in some other location.
Since the grasping is not as accurate as one would expect in many
cases, knowing the orientation of the object inside the gripper is
a very useful information to prepare the placing behavior.
This can be done either by in-hand manipulation, which usually
requires more fingers, or by adjusting the final target position
of the arm taking into consideration the object's orientation.


\begin{figure}
\centering
\includegraphics[width=\linewidth]{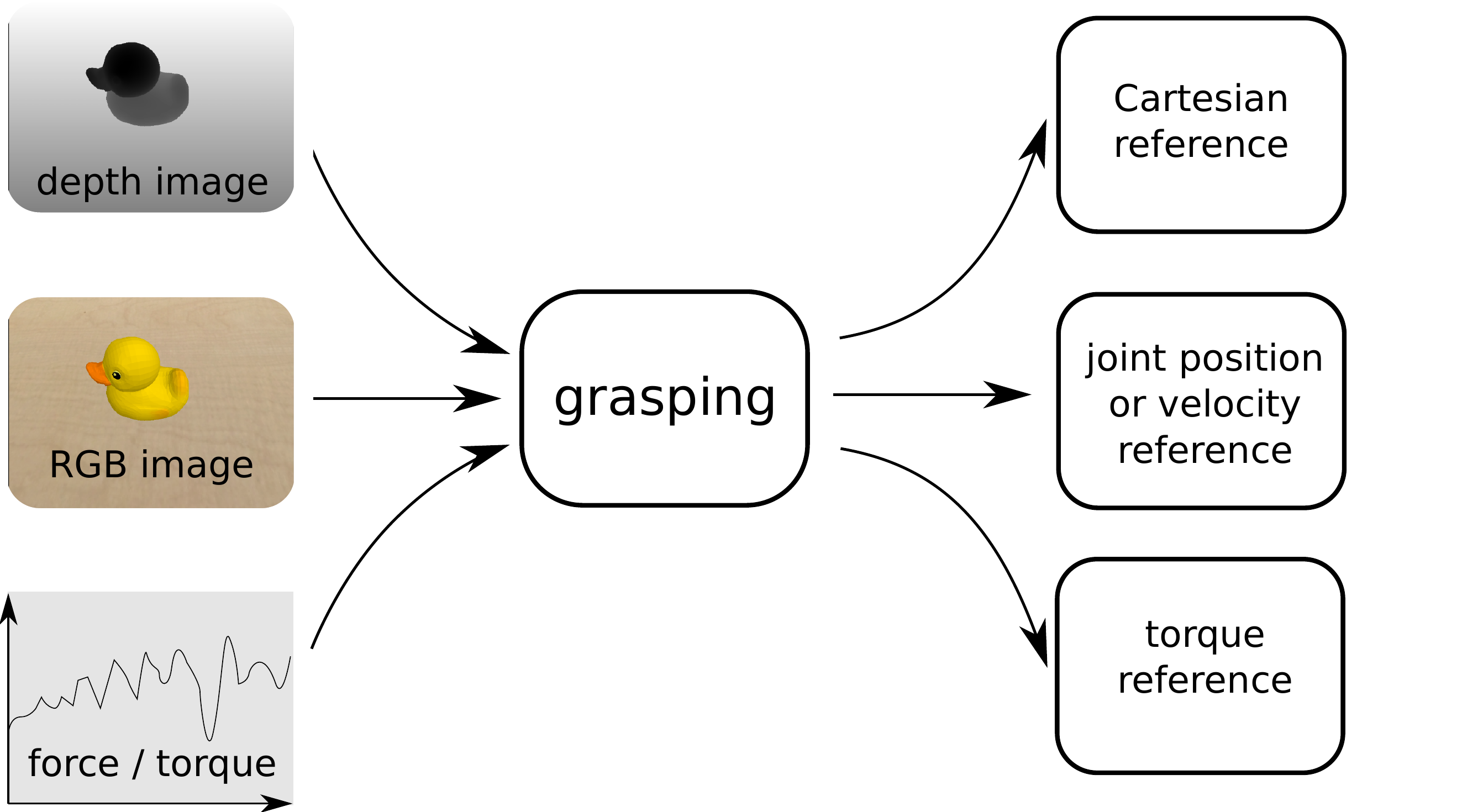}
\caption{\textbf{Learning grasping from sensory information.}
Exemplary sensor data that could be used to generate grasping
behaviors and possible outputs of a skill.\label{fig:grasping}}
\end{figure}

\paragraph{Pick and place:}
A skill that is very similar to grasping is pick and place.
Some works assume that picking the object is already solved
and learn only object placement \citep{Ijspeert2013,Finn2017},
others learn both pick and place in one policy
\citep{Stulp2012,Rahmatizadeh2016,Chebotar2017b}.
Some works only focus on movement execution \citep{Ijspeert2013},
others generalize from object features to trajectories \citep{Kroemer2017},
or even learn camera-based perception and action end-to-end
for one specific object \citep{Finn2017,Chebotar2017b}.
A very interesting work from \citet{Stulp2012}
considers the special case of this problem under uncertainty. It assumes a state
estimation approach to track the object's location which does not yield
perfect results. In addition, a sequence of movements is learned.
A variant of pick and place is placing coat hanger on a rack. \citet{Levine2016}
learned to perform this task end-to-end from camera images to motor torques.

The next level of difficulty for simple pick and
place tasks is placing objects precisely, for example, stacking boxes.
An interesting work shows that this can be learned even with
a low cost manipulator that has play in its joints and a wobbling base
\citep{Deisenroth2015}.
While this can be easily interpreted as noise from a machine learning
perspective, other methods usually fail without any informative prior
knowledge.
In their study, perception has
not been learned but continuous feedback from a vision system has been
used to generate appropriate action. \citet{Duan2017} tackle a more difficult
problem by learning a direct mapping from visual input to actions.
In their work, however, a more precise robotic system has been used.

\paragraph{In-hand manipulation:}
As objects cannot always be picked up in a specific configuration,
in-hand manipulation may be necessary to reposition the objects within a
robot's hand. In general, this is a dexterous manipulation skill that
requires a gripper with multiple fingers that can be driven individually.
\citet{Hoof2015} learn robot in-hand manipulation with unknown objects by
using a passively compliant hand with two fingers and exploiting tactile
feedback.
They investigate an in-hand object rolling task and learn a control policy that
generalizes to novel rollable cylindrical objects that differ in diameter,
surface texture and weight. In their work, dynamics and kinematics of the
compliant robot hand are unknown to the learning algorithm.

The hand used by \citet{Rajeswaran2018} has five fingers and has pneumatic
actuation. They consider the problem of learning in-hand rotation of
elongated objects with and without the use of a wrist joint under varying
initial conditions. The object can either be in the hand at the start of
the behavior or picked up and moved to the desired configuration.
Learning this skill is shown to be possible with only proprioceptive feedback.
This includes pressure measurements, positions, and velocities of each joint.

\cite{Andrychowicz2018} learn a very complex in-hand manipulation skill:
changing the orientation of a cube to any desired orientation in a
robotic hand with five fingers.
Two components are learned: a vision component that computes the object's pose
from three camera images from significantly different, fixed perspectives
and a policy component that uses the finger tip positions and the object pose
to generate motion commands for the fingers.
The finger tip positions are measured with a motion capture system
which unfortunately makes the learned skill in its current form not suitable
for a humanoid robot outside of the lab.

\paragraph{Tumbling / tilting an object:}
The challenge in quasi-static manipulation tasks like tumbling or tilting
objects from one face to another is to control the position of the
respective object over a period of time. \citet{Pollard2004} generalize
a object-tumbling skill to novel object sizes, shapes and
friction coefficients. \citet{Kroemer2017} further enhance the difficulty
by learning to tilt objects exactly around their defined pivotal corners.
This task requires a high accuracy during the whole skill execution because
the object's corner has to stay continuously in contact with the desired
pivot point.

\subsubsection{Deformable Objects (D)}

\paragraph{Knot tying and untying:}
Tying a knot is a behavior that is frequently required, for instance, during
surgical operations, in the household domain, for search and rescue, or
for sailing where threads or ropes are often used.
\citet{vandenBerg2010} demonstrate that a combination of behavior learning and
optimal control can be used to learn fast and smooth knot tying with two manipulators
consisting of 14 motors. This would be a particularly challenging task for planning
algorithms that would have to reason about a three-dimensional soft body.

Similarly, untangling ropes and untying knots is required in the very same domains as well as for technical applications in which cables unintentionally tangle up. 
\citet{wenHaoLui2013} learn to predict the rope configuration and use it to choose several actions from a predefined set to untangle the rope. 

\paragraph{Handling Garments:}

\citet{Corona2018} learn to handle garment, that is, arranging garment
from an unknown configuration to a reference configuration from which
further steps can be executed, for example, folding it or dressing a
person.
The difficult part is the prediction of suitable grasp points from
camera images. A bimanual setup has been used: one arm grasps a
garment and presents it to an RGB-D camera, the garment is recognized,
and two grasping points for the arms are identified to bring the
garment to a reference configuration. Jeans, T-shirts, jumpers, and
towels can be handled by the system.


\citet{Colome2018} learn to fold a polo shirt with two robotic arms. Each arm
has 7 DOF. Only trajectories for two arms are learned. An accurate model
of the polo shirt and its interaction with the grippers of the arms is not
available. The learned trajectories minimize wrinkles in the shirt and make
it look as close to a reference rectangle as possible.


\citet{Erickson2018} consider the problem of robot-assisted dressing:
while a human is holding his arm up and holds his posture strictly,
a PR2 robot pulls a hospital gown onto the arm of human.
Physical implications of actions on people are learned from simulation.
The learned model predicts forces on a person's body from the kinematic
configuration and executed actions. The model is combined with
model predictive control to solve the task. Hence, neither action, nor
perception are learned completely.

\subsubsection{Divisible Objects (E)}

\paragraph{Cutting:}
Cutting objects is a complex task as dynamics are induced during the process
of object cutting. Cutting tasks can be found in various domains. For example,
\citet{Lioutikov2016} consider the task of cutting vegetables in a kitchen
scenario. In their work, the movement is divided into multiple steps, and
afterwards executed autonomously as a sequence.
The learned behavior generalizes to changed cutting positions.
However, they do neither consider the required forces to cut the objects nor
the involved dynamics.
As a result, the cutting motion has to be executed multiple times to finally
slice the vegetable. Therefore, while \citet{Lioutikov2016} represent cutting
motions as discrete behaviors, they recommend to represent them as rhythmic
behaviors in future work.
The difficulty of food-cutting tasks is further exacerbated, if vegetables with
different stiffness and shape are evaluated. In this case, the (non-linear)
dynamics vary not only with time but also with different object types. As the
hand-designing of such dynamics models is infeasible, \citet{Lenz2015b} aim to
learn the prediction of these dynamics and the respective controllers directly
from a dataset of about 1500 cuts.
In the medical field, \citet{Thananjeyan2017} investigate surgical pattern
cutting of deformable tissue phantoms in the context of laparoscopic surgery.
As the task requires simultaneous tensioning and cutting, they learn a
tensioning policy which depends on the specific cutting trajectory and maps
the current state of the gauze to output a direction of pulling. Similar to the
work from \citet{Lenz2015b}, the dynamical deformation is difficult to observe
or to model analytically. Therefore, they directly learn the cutting policy in
an end-to-end fashion.

A similar task is peeling which has been learned by \citet{Medina2017}. It
is, however, modeled as a sequence of reaching, peeling and retracting.
Only with one arm the peeling motion for a zucchini has been learned while
another arm holds it.

\subsubsection{Movable Objects, Dynamic Behavior (F)}

\paragraph{Batting, throwing and kicking:}
For many games some sort of batting or throwing behavior is required,
for example, hockey \citep{Daniel2013,Chebotar2017,Rakicevic2017,Paraschos2018},
golf \citep{Maeda2016}, minigolf \citep{KhansariZadeh2012}, billiard
\citep{Atkeson1997,Pastor2011}, baseball \citep{Peters2005,Peters2008},
badminton \citep{Liu2013}, tennis \citep{Ijspeert2002}, table tennis
\citep{Kober2010,Muelling2011,Kober2012,Muelling2013}, tetherball
\citep{Daniel2012,Parisi2015}, darts \citep{Kober2012}, throwing
\citep{Gams2010,Ude2010,Kober2012,daSilva2014,Gutzeit2018}, and kicking
\citep{Boeckmann2016,Hester2010,Asada1996}.
These are very dynamic manipulation behaviors because momentum
from the end-effector has to be transferred to the manipulated object.
We can distinguish between settings where a specific goal has to be reached
by hitting or throwing an object directly
\citep{Chebotar2017,KhansariZadeh2012,Rakicevic2017,Paraschos2018,Gams2010,Ude2010,daSilva2014,Gutzeit2018}
or indirectly \citep{Daniel2013,Atkeson1997},
or the distance or velocity has to be maximized \citep{Pastor2011,Peters2005,Peters2008}.
Sometimes performing the motion was enough \citep{Maeda2016,Liu2013,Ijspeert2002,Daniel2012,Boeckmann2016}.
Winning the game was the goal in the case of tetherball \citep{Parisi2015},
or scoring a goal in the case of soccer \citep{Hester2010,Asada1996}.
An extension to the problem of hitting a specific goal is to hit a given
goal from a target space, for example, along a line \citep{KhansariZadeh2012},
from an area \citep{Kober2012,Gams2010,Ude2010,daSilva2014,Rakicevic2017,Gutzeit2018},
or from a discrete set of targets \citep{Kober2012}.
In some cases specialized machines have been used, for example,
\citet{Atkeson1997} use a simple billiard robot or \citet{Liu2013} use a
badminton robot with three DOF. In contrast, \citet{Pastor2011}
use a humanoid robot to play billiard or \citet{Muelling2013} use robotic arms
to play table tennis. In some works, only serve motions \citep{Liu2013} or
hitting static objects \citep{Peters2005,Hester2010} are learned, in other
works a moving object has to be hit
\citep{Muelling2013,Parisi2015}.
Perception and state estimation is not learned in any of the presented works,
hence, behaviors that rely on perception and state estimation of moving targets
\citep{Parisi2015,Muelling2013} can be considered as deliberative. Most of
these problems, however, have been solved without exterioceptive sensors.
Kicking a ball with a legged humanoid represents a particular challenge
because the robot has to keep balance. \citet{Boeckmann2016} execute a learned kick
with manually implemented balancing and \citet{Hester2010} learn to perform
a kick that avoids falling over while scoring a goal.
State estimation uncertainty and noise is an issue if perception is involved
in the skill although this has not been mentioned explicitly in the works
of \citet{Parisi2015,Muelling2013} in which state estimation methods have been
used.
Hence, we assume this has not been
considered to be a significant problem. Learning the perception part of
these behaviors has not been considered so far and would significantly
increase the difficulty of the problems.


\paragraph{More dynamic manipulation behaviors:}\label{Paragraph:OtherDynamicManipulationBehaviors}
In ball-in-a-cup, a ball is attached to a cup by a string. The goal is to
move the cup to catch the ball with it. A robot has to swing the ball up
and catch it. The movements of the ball are very sensitive to small
pertubations of the initial conditions or the trajectory of the end-effector
\citep{Kober2008}. Successful behaviors are learned so that they take into
account the ball position \citep{Kober2008,Kober2009} to compensate for
perturbations, however, the perception part is not learned in any of these
works. \citet{Kober2008} state that it is a hard motor learning task for
children.

Another remarkable work is published by \citet{Kormushev2010b}. The goal is
to flip a pancake with a frying pan. It is a dynamic task and the pancake is
susceptible to the influence of air flow which makes it very hard to predict
its trajectory.

\citet{Zhao2018} learn nunchaku flipping, which is a very dynamic behavior.
A nunchaku is a weapon that consists of two sticks that are connected by a chain.
A hand with haptic sensors and five fingers has been used.
\citet{Zhao2018} emphasize that the task requires compound
actions that have to be timed well, contact-rich interation with the
manipulated object, and handling an object with multiple parts of
different materials and rigidities.

\paragraph{Balancing:}
A typical balancing example which is often used as a sample problem is balancing an
inverted pendulum.
\citet{Marco2016,Doerr2017} investigate this problem in a real-world manipulation scenario by
utilizing a robotic arm with seven DOF to balance an inverted pendulum.
In their work, they learn parametrizations
of a PID controller or a linear-quadratic regulator (LQR), respectively,
while a motion capture system is used to track the angle of the balanced
pole.

\subsubsection{Granular Media and Fluids (G)}

\paragraph{Scooping:}
For humans, reasoning about fluids and granular media is no more difficult
than reasoning about rigid bodies. Not many researchers try to tackle these
problems with robots. \citet{Schenck2017} learn scoop and dump actions of
granular media.
Both are executed in sequence and they are encoded with nine parameters that
tell the robot where and how to scoop and where to dump the granular media.
The problem that is solved is to scoop pinto beans from one tray and dump
it to another tray to create a desired shape in the target tray.
A Gaussian-shaped pile and the letter ``G'' have been selected as target
shapes. The robot was allowed to execute 100 scoop and dump actions.
A depth camera is used to measure the current state of the granular media.
The part of the behavior that has been learned is a model that predicts
the effect of actions which will then be used to select good actions.


\paragraph{Pouring:}
An application which requires (weak) dynamical movements with moderate
precision is pouring liquids from a bottle into a cup.
Learning focuses on the generalization of
the movement to new goals (position of the cup \citep{Pastor2008}), changed
initial positions (position of the bottle \citep{Chi2017}), or different object
shapes and sizes \citep{Brandi2014,Tamosiunaite2011}. \citet{Tamosiunaite2011}
learn both, the shape of the trajectory and the goal position to generalize a
trajectory to a different bottle.
Similar to the pick-and-place applications detailed above, the elementary
pouring problem can also be extended to a pick-and-pour task
\citep{Caccavale2018,Chi2017}.
In contrast to the above mentioned works which acquire the pouring trajectories
from human demonstrations, robotic pouring behaviors can also be learned in an end-to-end
fashion directly from videos \citep{Sermanet2017}.

\subsubsection{Collision Avoidance (H)} \label{manipulation:collisionavoidance}
Robotic manipulation behaviors can result in collisions with the robot's
own body, other agents or the environment.
The latter is often termed obstacle avoidance, where the
obstacles can be both static or dynamic. While static objects in the
environment can be modeled well within a world model, dynamic obstacles
are often circumnavigated with reactive behaviors.
Both, collision and obstacle avoidance are important in real-world
manipulation scenarios.
\citet{Koert2016} learn adaptation of trajectories in case of unforeseen
static obstacles represented by a point cloud that has been obtained from
a depth camera.


\subsubsection{Miscellaneous (I)}
There are also some more unusual behaviors that have been learned but
we will not discuss them in detail.
Among these are archery \citep{Kormushev2010c}, which is similar
to throwing a ball or darts but does not involve an accelerating trajectory,
playing with the Astrojax toy \citep{Paraschos2018}, playing maracas
\citep{Paraschos2018}, drumming \citep{Ude2010}, and calligraphy
\citep{Omair2015}.


\subsection{Locomotion Behaviors}

The design of locomotion behaviors is a challenge that increases with the
kinematic complexity of the robot, its inherent stability,
and the terrain to be traversed.
Machine learning techniques can be used to provide solutions to
locomotion problems, even with fundamental principles of robot locomotion
not yet fully understood \citep{aguilar_review_2016}. 

Locomotion problems can be organized hierarchically based on the controlled
entities (single or multiple legs, joints of the robot body)
as shown in Figure~\ref{fig:behavior_hierarchy}.
On the lowest level, a PID controller may generate actuator
commands to control the joints of a robot leg or the motors of its wheels to
reach or maintain a certain position, velocity, or torque. By variation
of its parameters, a joint controller can achieve meaningful reactive
movements without knowledge of the kinematic structure.
As an example, each joint can independently compensate for internal friction
or a certain reflex can be triggered locally at joint level \citep{Kuehn2014}.
We exclude the level of joint control as it is only modifying a given
behavior generated on higher levels.
Single leg behaviors, such as the swing movement, can be defined in the
Cartesian space of the end-effector and thus require an inverse
kinematics and / or dynamics transferring the behavior's output into
joint space.
Behaviors that command the full body such as balancing or walking often
use other behaviors that only control single legs. High-level locomotion
behaviors concatenate, combine, and steer full-body behaviors. For example,
navigation behavior for a humanoid robot controls the goal of a walking
behavior. High-level behaviors could as well be controlled by other behaviors
or overall objectives.

\begin{figure*}[tb]
\centering
\includegraphics[width=0.7\textwidth]{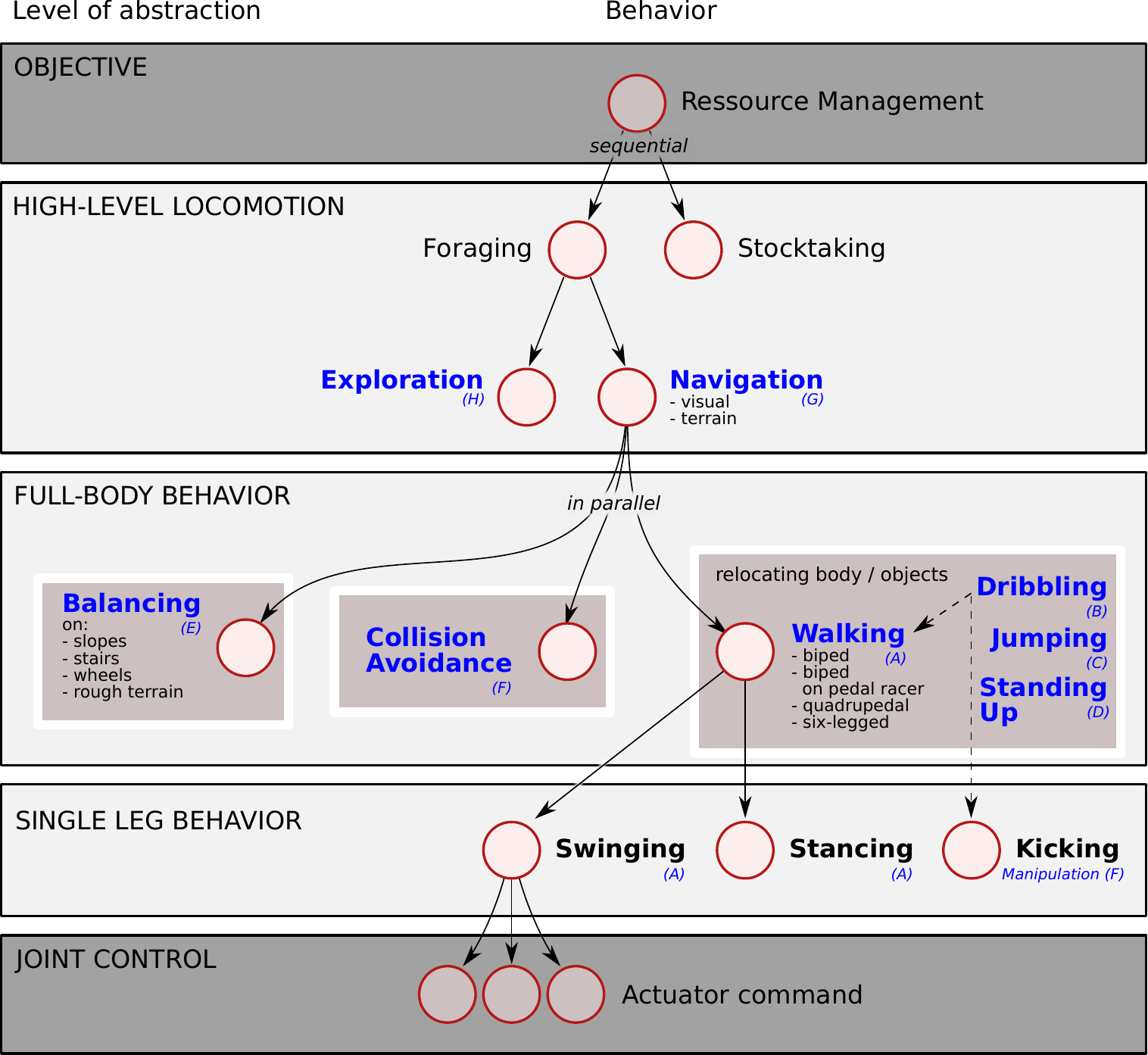}
\caption{
\textbf{Hierarchy of behaviors with focus on locomotion.}
Inspired by~\citet[page 49]{Arkin1998}. \label{fig:behavior_hierarchy}
For different levels of abstractions exemplary behaviors are presented.
Concrete movements of the body, a single extremity or joint are found on
lower levels in this hierarchy. While machine learning may be used on all
levels and intersections, this work focuses on behavior learning above the
level of joint control.}
\end{figure*}

\subsubsection{Walking (A)}
The prime example of the category locomotion is walking. Walking
is a very diverse robotic behavior learning problem. Its diversity stems on
the one hand from the variety of different walking machines:
six-legged \citep{Maes1990,Kirchner1997}, quadrupedal \citep{Kohl2004,Kolter2008,Birdwell2007,Kolter2009,Kalakrishnan2009,Zucker2011,Bartsch2016},
or biped systems \citep{Benbrahim1997,Matsubara2005,Geng2006,Kormushev2011b,Missura2015}
have been considered for this paper.
On the other hand, the problem formulation can be made
more difficult by requiring the system to walk up stairs \citep{Kolter2009} or
walk on irregular or rough terrain \citep{Kolter2008,Kalakrishnan2009,Zucker2011}.
In principle, the problems of walking as fast \citep{Kohl2004},
straight \citep{Birdwell2007}, energy-efficient
\citep{Kormushev2011b}, or stable \citep{Missura2015} as possible can be distinguished.
While six-legged and quadrupedal systems are stable enough to prevent falling
over in most situations and, hence, qualify for static behaviors, bipedal
systems are often unstable and it is a hard problem to prevent them from
falling over. Hence, bipedal walking can be considered a dynamic learning
problem. Walking is a rhythmic and active behavior. It is an elementary skill
that can be used in many application domains, however, walking robots are
in competition to wheeled robots which are much more energy-efficient
and precise in flat terrain.
While walking itself is a rhythmic behavior, precise foot placement is
usually a discrete behavior.
Precise foot placement is required for climbing stairs \citep{Kolter2009}
and walking on rough terrain \citep{Kolter2008,Kalakrishnan2009,Zucker2011}
on a lower level of behavior abstraction (see Figure~\ref{fig:behavior_hierarchy}).
Those behaviors also combine learning methods with other planning and control methods.
Bipedal robots are usually leaner than other walking machines and they are able
to move like humans and in the same environment, for example, go through very
narrow paths \citep{Benbrahim1997}. Because bipedal walking is not statically
stable per se, controllers have to compensate disturbances continuously.
Either static stability or dynamic stability can be the goal of a bipedal walk.
Often the problem
of learning bipedal walking is restricted by supporting structures to the
sagittal plane to simplify the balancing problem
\citep{Benbrahim1997,Matsubara2005,Geng2006} but not always
\citep{Kormushev2011b,Missura2015}. However, behaviors are often
prestructured to restrict and, hence, simplify the learning problem.
For example, \citet{Missura2015} only learn the balancing part of the
walk. Using sensory feedback is particularly important for bipedal walking.
Apart from proprioceptive sensors \citep{Matsubara2005}, ground contact
sensors have been used \citep{Geng2006}. Robustness to slightly irregular
surfaces and changes of the robots dynamics have also been considered
\citep{Matsubara2005} for bipedal walking.

A more difficult version of bipedal walking is riding a pedal racer.
In principle, it is comparable but it is crucial to exert a controlled force
on the pedals. Hence, \citet{Gams2014} use a 6-DOF force-torque sensor in
each foot of the bipedal robot to generate feedback to the learned
behavior.


\subsubsection{Dribbling (B)}
Walking or running while controlling a ball is called dribbling. It can be
used, for example, in basketball, handball, or soccer. \citet{Latzke2007}
learned dribbling for soccer with a humanoid toy robot by ``walking against
the ball''. The walking behavior is very simple because it only uses
three motors. The goal is to learn how to score a goal with dribbling,
starting from ten different initial ball positions at the middle of the
field. Only high-level control, that is, setting a walking direction has been
learned. Positions of the ball and the goal are obtained from a world model.


\subsubsection{Jumping (C)}
If the walking robot is too small and the terrain too rough, jumping is
sometimes necessary. \citet{Kolter2009} show that this can be used to climb
up large stairs with a small quadrupedal robot. With the same robot,
\citet{Theodorou2010} learn to jump across a gap by maximizing the distance
of the jump while jumping straight to prevent falling over. Unfortunately,
\citet{Theodorou2010} could not evaluate their approach on the real system.


\subsubsection{Standing Up (D)}
A stand-up behavior is important for any biped robot acting in the real world.
In general, the difficulty is that there exists
no static solution as there is no joint linking the robot to the ground.
For many robots, a robot-specific, pre-programmed stand-up movement
is used instead of acquiring the skill by learning.
However, \citet{Morimoto2001} learn a dynamical stand-up policy both in simulation
and on a real two joint robot. The robot (incrementally) learns a skill to
stand up dynamically by utilizing the momentum of its body mass.
An inclination sensor measures the current state of the system
and motor torques are produced by the learned motor skill.
The hierarchical learning architecture learns to generate postures by means of an upper level policy and the movements to achieve the next posture (sub-goals) by means of a lower level policy.

\subsubsection{Balancing (E)}
Keeping balance is a fundamental locomotion requirement and has been achieved with
various approaches by modifying different aspects of the motion.
For example, balancing a walking humanoid by modifying the gait
\citep{Missura2015}, using arm motions \citep{Kuindersma2011} or control motor
torques \citep{Vlassis2009} to balance a robot on two wheels.
Often behavior learning is combined with classical control approaches:
\citet{Kuindersma2011} use an existing balance controller for normal
balancing and only activate arm motions for postural recovery when
the inertial measurement unit (IMU) detects perturbations through impacts of
an external weight.

\subsubsection{Collision Avoidance (F)}
Learning collision avoidance seems to play a secondary role in manipulation
(see paragraph \emph{\hyperref[manipulation:collisionavoidance]{Manipulation: Collision Avoidance}}).
There are, however, many works in the context of locomotion, where
it is mainly related to navigation problems.
The publications discussed in this paragraph directly use images and vision
systems. They present learned reactive collision avoidance behaviors.
In the field of navigation, \citet{Tai2016b} learn a collision avoidance
strategy based on depth images in an indoor obstacle avoidance scenario.
They use a mobile, wheeled robot that learns to move in corridors
with a set of discrete actions. However, the robot only encounters static
obstacles. \citet{Loquercio2018} investigate a civilian drone flight
application. In their work, the drone learns to safely fly in the
streets of a city by mapping each single input image directly to a drone
steering angle and a collision probability to react to unforeseen obstacles.
The behavior for navigation and obstacle avoidance is trained for urban
environments from the viewpoint of bicycles and cars but can be generalized
to novel situations like indoor environments or high altitudes without
retraining. The outputs of the perception model are not directly used to
control the drone but converted to movement commands with fixed rules.
Similarly, \citet{Gandhi2017} also learn to navigate an unmanned
aerial vehicle while avoiding obstacles. They use negative experiences,
that is, a visual dataset of more than 11,500 crashes in various
environments with random objects, in conjunction with positive data to learn
to fly even in cluttered, dynamic indoor environments. The behavior is
learned end-to-end by taking camera images and outputting probabilities
of the motion commands go ``left'', ``right'', or ``straight''.
\citet{Kahn2017} learn uncertainty-aware collision avoidance, that is,
given a camera image and a sequence of controls the learned model will
output a collision probability together with an estimate of uncertainty.
The approach proceeds cautiously in unfamiliar environments and increases
velocity in areas of higher confidence. Model predictive control is used to
generate actions, while the cost model incorporates collision probability
and uncertainty. The approach has been tested with a quadrotor and an RC car.



\subsubsection{Navigation (G)}
Assuming the robotic system knows how to walk or drive, where should it move?
High-level locomotion behaviors like navigation and exploration are concerned
with local direction generation, for example,
navigation through complex natural environments \citep{Silver2010},
navigation to visually presented targets \citep{Zhu2017},
navigation to targets with known relative location \citep{Pfeiffer2016},
lane following \citep{Chuang2018},
reducing state estimation uncertainty in navigation \citep{Oswald2010} and
navigating to a target position \citep{Conn2007}. 
Most of the works discussed here are concerned with wheeled robots but
are in principle transferable to walking robots.
Classical navigation through natural terrain has been considered by
\citet{Silver2010}. They use planning to generate driving directions
but the generation of cost maps for the planner are learned.
The cost maps are generated based on perceptual data:
static data sources like satellite images or onboard sensors like
cameras and LiDAR.
\citet{Zhu2017} consider the problem of visual navigation:
actions in a 3D environment are predicted based on the current image from
the robot's camera and an image of the target.
The predicted actions result in a minimum path length to reach the goal.
They show that navigation to different targets in a scene can be learned
without retraining. The approach has been tested on a wheeled robot in an
office environment.
\citet{Pfeiffer2016} learn navigation to a given relative target location
end-to-end from 2D-laser range findings without a map. Steering commands
are directly generated by the learned behavior. The goal was to navigate
safely through obstacle-cluttered environments with a mobile platform.
A similar problem is to learn lane following from camera images end-to-end.
This has been done by \citet{Chuang2018}.
\citet{Oswald2010} consider the problem of navigation with a humanoid robot
that has noisy actuators and sensors. Motion commands are executed
more inaccurately with walking robots compared to wheeled robots and camera images are affected by
motion blur. A navigation behavior has to trade off quality of pose estimation
and walking speed. A vision-based pose estimation has been used and navigation
actions (forward, rotate left / right, stand still) for the robot have
been learned and take into consideration distance and angle to the goal
and pose uncertainty. The goal is to reach the destination reliably and
as fast as possible.
\citet{Conn2007} solve a classical grid-world navigation problem
in the real world. The laser scan data and orientation information is
used by the behavior to generate one of the commands stop, turn left,
turn right, or move forward.

As a side note, we would like to mention here that autonomous driving
behaviors for cars also fall into the category of navigation.
These behaviors can also be learned as shown by \citet{Chen2015,Bojarski2016}.
Because this topic is very broad and it is not of utmost importance for
humanoid robots, we will not further investigate it here.
The behaviors are often very specific for the domain, for example,
\citet{Bojarski2016} present an approach to learn lane and road following
and \citet{Chen2015} learn driving in a car racing game.


\subsubsection{Exploration (H)}
Exploration behaviors use (lower level) locomotion behaviors to gain knowledge
on the robot's environment.
\citet{Cocora2006} successfully transfer exploration behavior from other
environments to a new environment to find the entrance of an office.
The general problem that they try to solve is navigating to a room with an
unknown location. While searching for it, only labels for neighboring rooms are provided to the robot. The required exploration behavior is achieved by learning an abstract navigation policy choosing actions based on the provided local knowledge. 
\citet{Kollar2008} learn an exploration behavior for an unknown environment
to maximize the accuracy of a map that is built with simultaneous localization
and mapping (SLAM).

A special case of exploration behaviors are sampling routines aimed at
acquiring relevant sensory input often referred to as active sensing or
active perception. 
\citet{chen_active_2011} state that ``active perception mostly encourages the idea
of moving a sensor to constrain interpretation of its environment''
For example, a camera usually has a limited field of view, thus,
the goal of an active sensing behavior is to move the part of the
robot to which the camera is attached (or the whole robot) to reduce
uncertainty about the scene.

\citet{Kwok2004} demonstrate how active sensing can be learned
in the domain of robotic soccer: a quadrupedal robot has to
determine its own location, the location of the ball, and the location of opponents
on a soccer field with a camera to finally score a goal.
The behavior considers the current estimate of the world state
and its uncertainty from the state estimation component. It
generates head motions to change the camera position. The robot
is trained to score a goal. The active sensing behavior is
executed while the normal soccer behavior is running.



\subsection{Other Behaviors}

Some behaviors cannot generally or not at all be classified as locomotion or
manipulation. We will discuss these behaviors in this section.

\subsubsection{Human-robot Interaction}
Human-robot interaction has become a feasible application through safe,
compliant robot control and design. Robots can come into physical contact with
humans in these scenarios.
Robots that assist humans with their tasks are particularly appealing in the
household and manufacturing domains.
They can hold objects for a human \citep{Ewerton2015}, hand over objects to a
human \citep{Ewerton2015,Maeda2017}, assist a human in putting on a shoe
\citep{Canal2018}, lift \citep{Evrard2009} or carry objects in collaboration
with a human \citep{Berger2012,Rozo2015},
or drill screws placed by a human \citep{Nikolaidis2013},
hence, show collaborative behavior.
They can even interact socially with humans, for example, by giving a high five
\citep{Amor2014} or shaking hands \citep{Huang2018}.
These behaviors are dynamic because they have to be synchronized with the human.
Challenging tasks are the recognition of the human's
intention and acting accordingly. Some authors focus on the intention
recognition: \citet{Amor2014} only consider the problem of recognizing one
interaction scenario by observing the human's motion, whereas
\citet{Ewerton2015,Maeda2017} consider the problem of distinguishing between
several possible interaction scenarios.
In these works, only marker-based motion capture systems have been
used to provide motion data from the human counterpart.
The presented behaviors are active, discrete manipulation behaviors and
perception has not been considered.
What makes carrying special is that it is a collaborative behavior
which requires continuous observation of the co-worker's state and intention
because both agents are indirectly physically connected during the whole
activity.
Carrying an object in collaboration of a robotic arm and a human might require
exerting a specific force on the object, and therefore, a method to measure
the forces. \citet{Rozo2015} use a 6-axis force/torque sensor for this.
In their application, the object can only be carried if both agents apply
a force in opposite directions.
In contrast, \citet{Berger2012} consider collaborative carrying as a whole
body problem with a humanoid. They adapt the walking direction of a robot
according the movement of its human counterpart. Deviations from learned
expected movements are recognized and the motion is adjusted accordingly.
In this case only part of the perception is learned.
Carrying behavior is often done with the robot following the human leader.
They can be considered passive.
The similar problem of lifting an object in collaboration has been considered
by \citet{Evrard2009}. They additionally learn to recognize if the robot
should take the leader or follower role during task execution.
Hence, the learned behavior can be both active or passive.
\citet{Canal2018} provide an example of a deliberative system,
where low-level actions have been learned and high-level symbolic
planning is used to organize communication and interaction with
a human.
They study the application of assisting a human in putting on a shoe.
The social acceptance of robots is an important aspect for future
robots interacting with humans. One of the key factors in this context are
natural motions, that is, the robot should not only reach a certain pose of
the end-effector but also execute the motion in a human-like manner.
To achieve this, \citet{Huang2018} present a hybrid space learning approach
that learns and adapts robot trajectories in Cartesian and joint space
simultaneously while taking into account various constraints in both spaces.
They evaluate their approach on a humanoid robot in a hand-shaking task,
consisting of a discrete reaching and a rhythmic waving motion, and adapt
the movement to different areas for shaking hands.
\citet{Nikolaidis2013} present results in a simplied human-robot collaboration
scenario. The scenario should model the human-robot interaction challenges
that occur in a hybrid team of a human and a robot that has to drill screws.
The human has to place screws and the robot drills them. Although in the
real world scenario there are no real screws and not a real drill, the
robot learns to execute its motions in an order favored by the human.
The problem of perceiving the human's current state is simplified by using
a motion capture system.



\subsubsection{Behavior Sequences}
The very specific task of unscrewing a light bulb is a good example for
sequential tasks that need to be decomposed into smaller subtasks to
achieve the overall objective. \citet{Manschitz2016} infer an unknown
number of such subtasks automatically from demonstrations of the overall
task and learn how to sequence the subtasks in order to reproduce the
complete task. In their work, the taught task sequence consists of
approaching the light bulb, closing the end-effector, unscrewing the bulb
by rotating the wrist stepwise (after each turning, the fingers are opened
and the wrist is rotated back), pulling the light bulb out of the holder and
finally putting it into a box.

Besides the applications of pouring, cutting and wiping, another typical
kitchen task is cooking (see also pancake flipping described in paragraph
\emph{\hyperref[Paragraph:OtherDynamicManipulationBehaviors]{
More dynamic manipulation behaviors})} or, more specifically, food preparation.
The preparation of food requires very structured behaviors with a fixed
chronological order of actions. Therefore, the complete task has to be
segmented into smaller sub-tasks. The order of these sub-tasks is typically
managed by a higher-level monitoring system. \citet{Caccavale2018}
picked the tasks of coffee and tea preparation to present their work on learning
the execution of structured cooperative tasks from human demonstrations
(respectively, though only in simulation, \citet{Caccavale2017} investigated pizza preparation).
A similar approach was presented by \citet{Figueroa2016}
on pizza dough rolling task with the goal to achieve a desired size and
shape of the pizza dough.

\subsubsection{Soccer Skills}
Soccer is one of the most extensively studied games in robotics.
Besides walking, dribbling and kicking, more high-level skills have been
learned with simpler robotic systems or in simulation. For example,
\citet{Mueller2007} learn ball interception on a wheeled robot
with known poses and velocities of the ball and the robot,
\citet{Riedmiller2009} learn an aggressive defense behavior also
based on these information and the pose and velocity of the opponent
but only in simulation, \citet{Riedmiller2007} learn cooperative team behavior
also in simulation.
Another example of a low-level behavior that has been learned for robotic
soccer is capturing a ball with the chin of a dog-like robot
\citep{Fidelman2004}.


\subsubsection{Adaptation to Defects}
A kind of learned behavior that does not fit into any category because
it is more general and can be used in combination with any underlying behavior
is presented by \citet{Cully2015}. The robot learned to adapt to defects.
A walking behavior of a six-legged robot as well as pick and place with a
manipulator with redundant joints have been considered.



\section{Discussion}
\label{s:discussion}


While we scanned the presented works, we made several interesting observations
that we will summarize in this section. Some statements certainly
depend on the machine learning method that is used, which we will indicate,
but most of our statements apply universally.

\subsection{What Makes the Domain Difficult?}

Learning on physical robots is difficult. There are numerous reasons why
much more machine learning is focused on only perception or is done in
artificial environments, for example, physical simulations.
We will summarize them here.

Robotic behaviors cannot be executed indefinitely often.
Robots suffer from wear and tear and hardware is often expensive
\citep{Kober2013}.
Robots can break. Robots can break things.
Robots require maintenance, for example,
battery changes and hardware repairs \citep{Kohl2004}.
Training data is often sparse. Learning methods must be effective with
small datasets \citep{Kohl2004}.
The main reason why human supervision is usually required is that many
behaviors require physical contact between robot and environment.
Hence, imperfect behavior might break either the robot or the environment
\citep{Conn2007,Englert2018}.
Robots change their properties over time.
Reasons can be wear or changing temperatures \citep{Kober2013}.
Behaviors cannot be executed faster than real time.
There is no way to speed this up like in simulations \citep{Fidelman2004}
besides adding more robots which require more maintenance work.
Simulation is difficult.
Dynamics of many robots and their environments are very complex and are
difficult to model.
\citet{Kohl2004} write
``robots are inherently situated in an unstructured environment with
unpredictable sensor and actuator noise, namely the real world.''
Curse of dimensionality is an issue.
Humanoid robots can have as many as forty or more state space dimensions
\citep{Morimoto2001}.
Behaviors have to be able to deal with partial observability,
uncertainty, and noise \citep{Kober2013}.
They are also often hard to reproduce \citep{Kober2013}.

Learning behaviors for robots in the real world is difficult for all those
reasons. Some of them can be mitigated in laboratory conditions but
this domain is still one of the hardest for todays machine learning
algorithms.

\subsection{When Should Behaviors Be Learned?}

One of the main questions that we would like to answer with this article
is which behaviors we should learn given the availability
of alternative approaches and difficulties applying machine learning to
real robotic systems. It is often intuitively clear to
machine learning and robotics researchers but the intuition is often
not underpinned by scientific evidence. The field is so diverse
that it is easy to miss something.

We see several strengths of learned behaviors that
have been mentioned quite often:
\begin{itemize}
\item Handling uncertainty and noise.
\item Dealing with inaccurate or non-existing models.
\item Learning can be better than hand-crafted solutions.
\item They are easier to implement.
\item They are often simple, sufficient or optimal heuristics.
\end{itemize}
We will back up these findings with sources in the following paragraphs.
Machine learning is also considered to be the direction
to real artificial intelligence or as \citet{Asada1996} put it:
``The ultimate goal of AI and Robotics is to realize autonomous
agents that organize their own internal structure in order to behave
adequately with respect to their goals and the world. That is, they
learn.''

\begin{figure}
\centering
\includegraphics[width=0.6\textwidth]{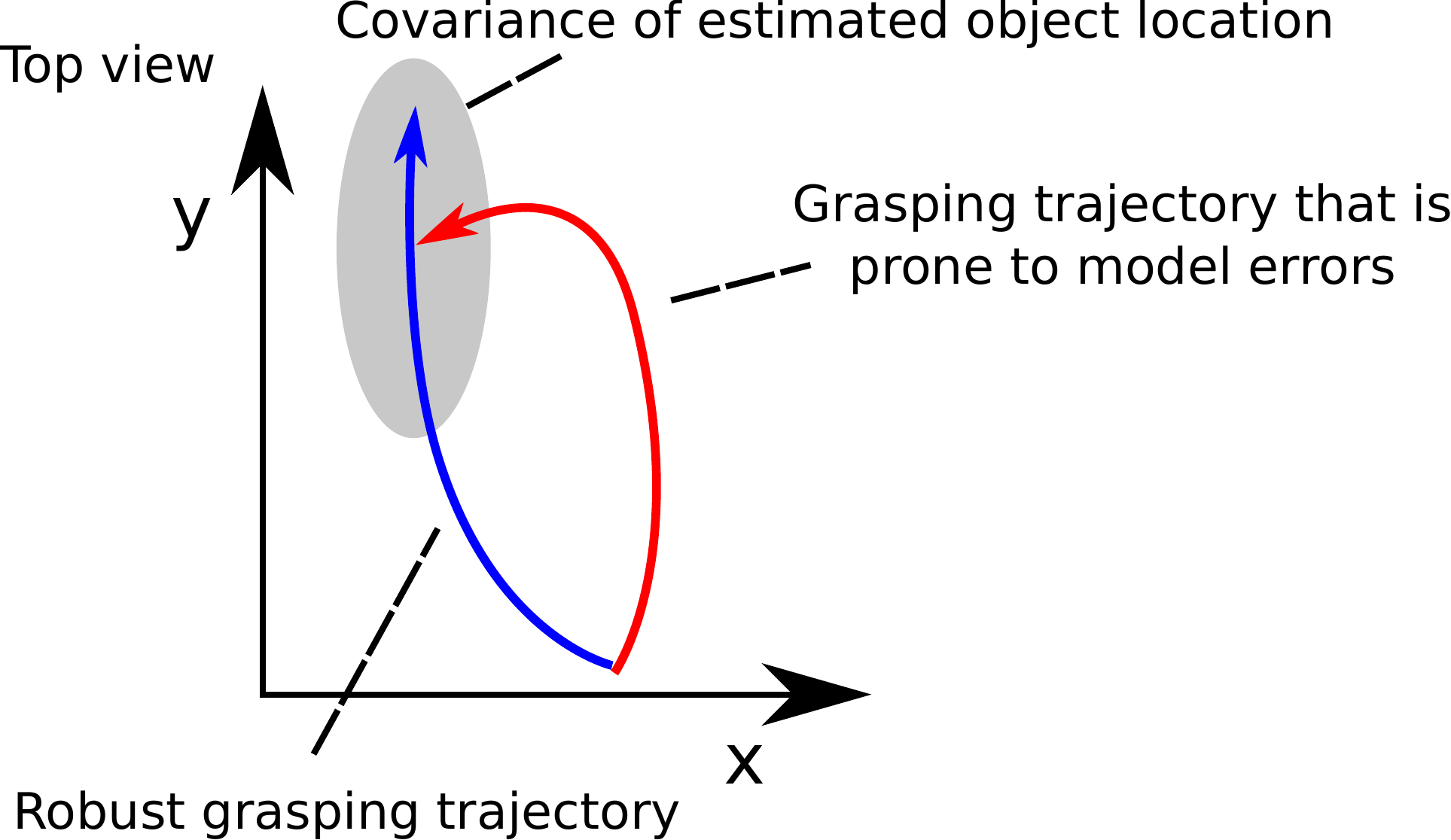}
\caption{\textbf{Sketch of a robust grasping trajectory from top view.}
The ellipse indicates the uncertainty of the objects estimated
position. A grasp that moves along the axis of highest variance
of the estimate (blue trajectory) will succeed with a higher probability than
a grasp that moves along the axis of lowest variance (red trajectory).
\label{fig:robustness}}
\end{figure}

Uncertainty and noise are two predominant problems in robotics.
Sensors and actuactors undoubtedly have to suffer from noise.
Noise, from the perspective of a robot, is part of nature and
it is an intrinsic property of these devices.
\citet{Mason2012} points out that uncertainty played a central
role in robotics research since its beginning.
Information about the world is usually incomplete and knowledge
is not certain.
This is the reason why probabilistic methods (see, for example,
\citet{Thrun2005}) are so popular in the robotics community.
\citet{Stulp2011,Stulp2012} show that state estimation uncertainty in a
pick and place problem can be compensated with an adapted motion.
We illustrate how a compensatory motion can address the problem
of state estimation uncertainty in Figure \ref{fig:robustness}.
An example of incomplete information is presented by
\citet{Levine2017}, where just a single RGB camera is used to
learn grasping end-to-end. The distance and the three-dimensional
structure of objects cannot be inferred from only one camera.
However, objects are in the same distance to the robot when
they are at the same position in the image. Hence, the system
must implicitly learn the objects' distance.
\citet{Laursen2018} explicitly design a method to help users in
creating robust and uncertainty-tolerant trajectories for assembly
operations which have previously been defined in simulation.
\citet{Deisenroth2015} use a low-cost robotic manipulator and
show that their method can compensate for actuator noise.
\citet{Carrera2012} state that learning offers the adaptability
and robustness that is required to solve their problem of turning
a valve.
\citet{Kober2008} learn a coupling of perception and action
to handle perturbations of trajectories.
\citet{Gullapalli1994} learn peg-in-a-hole insertion. They have
sensor noise in position encoders and in a wrist force sensor
and demonstrate that reinforcement learning can be used to generate
robust insertion behavior.
\citet{Johns2016} consider the problem of grasp pose prediction
and state that
``issuing commands to align a robot gripper with that precise pose
is highly challenging in practice, due to the uncertainty in gripper
pose which can arise from noisy measurements from joint encoders,
deformation of kinematic links, and inaccurate calibration between
the camera and the robot.'' They develop a method that explicitly
addresses these uncertainties.
Finally, \citet{Oswald2010} state that execution of motion commands
is noisy on a humanoid robot due to backlash in joints and foot slippage
and pose estimation during walking is more difficult because of motion blur.
They explicitly learn a high-level navigation behavior that reduces
pose estimation uncertainty that arises from the noise.

When there is no model of the robot or the world or existing models
are too inaccurate, machine learning can compensate for that.
This has been shown in the context of dynamic behaviors. It is hard
to model dynamics correctly but it is often not required. For example,
\citet{Muelling2013} use a state estimation to predict ball trajectories
in table tennis but neglected the spin of the ball.
\citet{Parisi2015} use a simplified model of the forward dynamics of
a robotic arm with springs. The learned behavior was able to work with
the simplified model.
\citet{Kormushev2011b} consider the problem of energy minimization in
a walking behavior that is used with a robot that has springs in its
legs. They claim that it is nearly impossible to solve the problem
analytically
``due to the difficulty in modeling accurately the properties of the
springs, the dynamics of the whole robot and various nonlinearities,
such as stiction.''
In general, soft bodies and soft-body dynamics are difficult to model
but that would be required, for example, for cutting and knot tying
behaviors.
\citet{Englert2018} write that a
``main issue is that the external degrees of freedom can only be
manipulated through contacts, which are difficult to plan since a precise
and detailed physical interaction model is often not available.
This issue motivates the use of learning methods for manipulation skills
that allow robots to learn how to manipulate the unknown environment.''
\citet{Colome2018} state that for problems that involve manipulation of
non-rigid objects accurate models are usually not available. Hence,
they use machine learning to solve the task of folding a polo shirt.

Direct comparisons of machine learning and hand-crafted approaches
have been done by \citet{Kohl2004,Kwok2004,Kober2008,Parisi2015}.
These works show that learning is able to yield better behaviors
than model-based or hand-tuned solutions. However, this result has
to be read carefully because it is certainly subject to publication
bias.
To our knowledge, there is almost no publication in which machine learning
for robotic behaviors and another method are compared with the result
that machine learning is worse.
Only \citet{Bargsten2016} compare machine learning with dynamic model
identification to learn a model of inverse dynamics with the result
that the machine learning method is worse because it does not generalize
well. Although it has to be noted that the dynamic model identification
is also a data-driven method with incorporated physical prior knowledge.
It is also not directly related to our survey because we excluded
low-level control.

Learning approaches are often easier to implement because they are
often general approaches and do not require problem-specific models.
Sometimes it is easier to specify the problem and not the solution.
A reward for reinforcement learning, for example, can encode the
problem specification. Examples of problems where it is easy to
define the reward are walking as fast or straight as possible,
jumping as far as possible, throwing as close to a target as possible,
or grasping: we could apply random perturbations after the grasp and
measure if the gripper still holds the object.
While ``walk as fast as possible'' alone might not be a sufficient
reward function, additional components of the reward function are
usually intuitive and part of the problem specification: we can
penalize behaviors that let the robot fall down or exert high
forces on parts of the robot.
\citet{Kormushev2010b} also made an interesting observation:
they found that the solution to the pancake flipping problem that has
been discovered by learning contains an unexpected compliant catching
behavior in the end of the movement.
This prevents the pancake from bouncing off the pan.
They conclude ``such undesigned discoveries made by the RL algorithm
highlight its important role for achieving adaptable and flexible robots''.
Imitation learning is another method that is particularly easy to use
from an end users perspective. It enables users to teach robots new
behaviors without requiring expert knowledge or programming skills
\citep{Alizadeh2014}.
We do not want to deny that tuning hyperparameters of a machine
learning algorithm is a complex task and requires expert knowledge,
but \citet{Parisi2015} found that tuning hyperparameters can be less
time intensive than building a mathematical model for a given task.
\citet{Amor2014} justify the use of machine learning in the context
of human-robot interaction:
``programming robots for such interaction scenarios is notoriously
hard, as it is difficult to foresee many possible actions and responses
of the human counterpart''.
\citet{Matsubara2005} learn a walking behavior and point out the
drawback of classical, model-based approaches. These require
precise modeling of the dynamics of the robot and the environment.
\citet{Fidelman2004} state that their paper
``is concerned with enabling a robot to learn high-level goal-oriented
behaviors. Coding these behaviors by hand can be time-consuming, and it
often leads to brittle solutions that need to be revised whenever the
environment changes or the low-level skills that comprise the behavior
are refined.''
\citet{Levine2017} start with the assumption that
``incorporating complex sensory inputs such as vision directly into a
feedback controller is exceedingly challenging'' and show with their
approach that learning complex emergent behavior can be done without
much prior knowledge.
Considering more the long-term perspectives of robotics and artificial
intelligence, the following two works are relevant.
\citet{Cully2015} consider the problem of adapting to hardware defects,
similar to injuries of animals. They found that
``while animals can quickly adapt to a wide variety of injuries, current
robots cannot 'think outside the box' to find a compensatory behavior when
damaged: they are limited to their pre-specified self-sensing abilities, can
diagnose only anticipated failure modes, and require a pre-programmed
contingency plan for every type of potential damage, an impracticality
for complex robots.''
\citet{Kirchner1997} considers the problem of an autonomous robot
that adapts its behavior online:
``if we face the general problem to program real robots to achieve goals
in real world domains, then, sooner or later, we will surely be confronted
with problems for which a solution is not at hand and probably can not even
be formulated off-line. In other words there are situations that the robot
might encounter during interaction with the real world, that we are not able
to foresee and we are therefore unable to precompile an appropriate set of
reactions for it. Yet, the robot needs to find the set of reactions by
itself. For this, learning is a necessity for real world robots.''

Before we elaborate on the the last point, we will draw an analogy
to behaviors of biological systems.
Most behavior learning algorithms that have been used in the works that have
been presented here do not guarantee optimality. We can consider the learned
behaviors to be heuristics.
Heuristics are often computationally efficient. That, however, does not make
them second-best strategies.
In real world situations, where an agent is embodied in a physical system with
sensors and actuators with noise and uncertainty, heuristics often yield useful
behaviors.
An often mentioned example for heuristic behavior is the gaze heuristic that
is used to catch a ball that is high up in the air \citep{Gigerenzer2009}:
``Fix your gaze on the ball, start running, and adjust your running speed so
that the angle of gaze remains constant.'' The player will be at the position
where the ball comes down. Other variables can be ignored, for example,
distance, velocity, and spin of the ball, air resistance, and speed and
direction of the wind.
\citet{Gigerenzer2008} explains why heuristics are useful in the case of
human behavior. These arguments are also applicable in the case of robotic
behaviors.
An optimal solution to a real-world problem is often computationally
intractable, for example, NP-hard or so ill-defined that we do not know
exactly what we should optimize for.
In addition, real-world problems demand for robustness of behaviors.
More information and computation is not always better
according to \citet{Gigerenzer2008}.
Reasoning often results in less successful behavior because of errors in
the model. Robustness sometimes even requires to
ignore or forget information.
From the papers that we read about learning robotic behaviors, the following
publications back up these statements.
\citet{vandenBerg2010} consider the problem of cutting, which would be hard
to model completely but has simple solutions.
\citet{Benbrahim1997} state: ``The fact that walking is most of the time done
unconsciously suggests that maybe it does not require constant heavy computing
in normal walking conditions.''
\citet{Kuindersma2011} learn balancing behaviors with arm motions and point out:
``This general problem also has several attributes that make it interesting
from a machine learning perspective: expensive evaluations, nonlinearity,
stochasticity, and high-dimensionality. In our experiments, a low-dimensional
policy space was identified \ldots''.


We will conclude with another view on the question why
machine learning should be used.
More than two decades ago, \citet{Thrun1995} already tried to answer the same
question.
They distinguish between model-based approaches (with a model
of the robot and the world) and learning. In a way we can consider every
approach that does not use machine learning to be model-based because it
either uses an explicit model (for example, planning, reasoning, or optimal
control) or an implicit model (for example, behavior definitions with
finite state machines or hard-coded motions).
Learned behaviors also build models but learned models directly encode
real experience.
\citet{Thrun1995} identify four bottlenecks of model-based methods.
There is a \textbf{knowledge bottleneck}:
knowledge has to be provided by a human. While this is not totally
accurate anymore because robots are, for example, able to build detailed
maps of their environment on their own, this is still an issue because a
programmer still has to define how the data is interpreted:
what is rigid and what is soft, which objects are movable and which are fixed?
There is an \textbf{engineering bottleneck}: it requires a lot of time to
implement and generate these explicit models.
For example, realistic modeling and physics simulation of soft bodies,
divisible bodies, deformable objects, fluids, or granular media are
still difficult.
There is a \textbf{tractability bottleneck}: many realistic problems are
computationally complex or even intractable which results in slow
responses. For example, \citet{Kuindersma2016} report times of
1.5 or 10 minutes to plan simple jumping motions.
There is a \textbf{precision bottleneck}: the robot must be able to
execute plans accurately enough. This is still an issue and is becoming
more relevant with flexible and compliant robots.

While all of the mentioned points are still valid, some of them
also apply to state-of-the-art machine learning. The knowledge
bottleneck is an issue if pre-structured policies or models are
used, for example, dynamical movement primitives \citep{Ijspeert2013}.
The tractability bottleneck has a counterpart in machine learning:
a lot of experience might be required.
As we have seen, simple heuristics are often sufficient,
which means that neither pre-structuring or restricting the policies
or models necessarily results in bad performance, nor will learning
require much data.
The precision bottleneck is related to the reality gap \citep{Jakobi1995}
that is a problem if behaviors are learned in simulation and
transferred to real systems. For example, \citet{Kwok2004} report this
problem.

\subsection{An Analogy: Shifting from Deliberative to Reactive Behaviors}

An often quoted statement from \citet[page 61]{Whitehead1911} is the following:
``It is a profoundly erroneous truism ... that we should cultivate the habit
of thinking of what we are doing. The precise opposite is the case.
Civilization advances by extending the number of important operations which
we can perform without thinking about them.''
Skilled human behavior is trained and repeated often.
Such a learned behavior is good because we do not waste many computational
resources. We are able to execute it fast and precisely.
\citet[pp. 100-101]{Norman2013} states:
``Conscious thinking takes time and mental resources. Well-learned skills
bypass the need for conscious oversight and control: conscious control is
only required for initial learning and for dealing with unexpected situations.
Continual practice automates the action cycle, minimizing the amount of
conscious thinking and problem-solving required to act. Most expert, skilled
behavior works this way, whether it is playing tennis or a musical instrument,
or doing mathematics and science. Experts minimize the need for conscious
reasoning.''
In other words \citep[page 2]{Shadmehr2005}:
``motor learning matters because it allows you to act while directing your
attention and intellect toward other matters. Imagine that you needed to
attend to all of the routine aspects of your reaching or pointing movements.
Motor learning provides you with freedom from such a life.''
Exactly the same statement could be made for robotic behaviors.
Learning individual skills also simplifies reasoning and planning because
planning can take place purely on a high level and solve the problem of
combining individual skills.

An argument in favor of learning robotic behaviors is this analogy to
well-learned human behavior.
As we have seen, learned behaviors are mostly reactive behaviors or heuristics.
This is the precise opposite of the very useful combination of mapping, state estimation,
and planning which we categorize as deliberative behavior. While state
estimation and planning works without previous interaction with the environment,
learned behaviors can be faster and can have a higher performance if
enough data is available or trials are allowed. While deliberative behavior can
be a safe first solution, it can be replaced by learned and reactive behaviors.
This is actually very similar to what humans do.

In summary, there is an analogy between humans and robots:
learned behavior can perform better while requiring less computational
resources in comparison to high-level reasoning in certain problem domains.

\subsection{When Should Behaviors Not Be Learned?}

Imagine you are a robot and you are in a critical situation that you have
never seen before. \citet{Dismukes2015} have an advice for you:
``identify and analyze decision options'' and ``step back mentally from the
moment-to-moment demands ... to establish a high-level ... mental model that
guides actions''.
Oh, you learned all of your behaviors end-to-end and you do not know how to
build a high-level mental model? Tough luck!

Not everything should be learned.
Learning in robotics often aims at reproducing the quality of human behavior
that cannot be reached by conventional approaches.
Humans are much better than robots at many tasks that require interpreting
complex sensory data, involve noise and uncertainty, and fast and dynamic
behavior.
They are the best examples of a learning, physical agent that we have
seen so far.
It is probably hard to achieve better results than a human if we try to use
the same design principles for robots.
Also humans make errors all the time and the frequency of errors can even
increase under external factors like stress
\citep{Dismukes2015}. While we do not think that robots are prone to stress,
we think that in learned robotic behaviors often unexpected failures might
occur.
A robot might encounter a situation that does not occur in the training set
(``distributional shift'', see \citet{Amodei2016}) or the agent learns
continuously which means that it also forgets.
Sometimes it makes sense to rely on logical reasoning and model-based
approaches. Ironically, \citet{Dismukes2015} propose the same for humans to
reduce errors under stress. It is the quoted advice from the previous
paragraph.


If a precise model of the world is available, planning and optimal control
often generate new behaviors faster and do not require physical interaction
with the real world before they provide a solution.
For example, collision avoidance based on distance sensors and planning
or reactive behaviors can be close to perfect so that it is applicable in
industrial scenarios \citep{deGeaFernandez2017}.
If collision avoidance is learned, there is no guarantee for safety.
Particularly, there will be no safe collision avoidance during the learning
phase, in which imperfect behaviors will be explored on the real system.
\citet{Tassa2012} show that, even if the model is not accurate,
model-predictive control (MPC; online trajectory optimization) with
a finite horizon can be used to generate intelligent and robust get-up
and balancing behaviors.
It has to be noted though, that optimal control and reinforcement learning
are related \citep{Sutton1992}. In this article we make the distinction
between reinforcement learning that needs experience and optimal control
that needs a model. Machine learning and optimal control
can be combined \citep{Levine2016,Erickson2018}.

Learning systems are typically not good at repetitive tasks and tasks that
demand for high precision, for example, tasks that have to be executed in a
factory. If the same car has to be produced several thousand times in precisely
the same way, it is worth the effort to let a human design the process step by
step. In a lot of situations it is even better to build specialized machines
instead of using robots. Robots and behavior learning only is required
if the system is confronted with changing requirements and environments.

Coordination of behaviors is a rather difficult task for machine learning
at the moment. Whole-body control \citep{Sentis2006} is quite successful in
this domain. It allows to prioritize tasks and solves everything online
in a high frequency on the system. If, for example, an existing walking
and object manipulation behavior should be combined so that the robot
keeps its balance, whole-body control is the method of choice.
Whole-body control is effective because it uses domain-specific knowledge:
the Jacobian of the robot. In order to exhibit similar behavior, a learned
behavior would implicitly have to approximate the Jacobian.
However, configuring whole-body control is challenging. Weighting and
prioritizing subtasks such that the result ``solves the task'' is
a difficult, manual task.

Perception for dynamic problems is challenging at the moment.
It can be learned for static behaviors like grasping \citep{Levine2017}
or visual servoing \citep{Levine2016} but it is nearly impossible at
the moment to learn a catching behavior for a ball end-to-end
because the learned model has to solve difficult perception, tracking,
and prediction problems while it must respond very fast.
\citet{Birbach2011} impressively show how computer vision and state estimation
can be used to track ball trajectories with an error of 1.5 cm in the predicted
catch point. The perception takes about 25 ms and tracking about 10 ms per
step. A ball catch rate of 80 \% has been reached on a humanoid upper body.

Learned behavior can show emergent properties. While this is sometimes
good, for example, in the case of the pancake flipping task
\citep{Kormushev2010b}, it can also be disastrous. For example, in
reinforcement learning or similar disciplines learning algorithms often
exploit ill-posed problem definitions. This is called ``reward hacking''
\citep[pages 7--11]{Amodei2016} and it is not necessarily immediately
visible. This problem can be particularly challenging if the behavior
should be used in a variety of different contexts and environments.


Interestingly, ``playing soccer'' is one of the most complex high-level
behaviors that robots are able to perform today and it is not learned.
On the contrary, it is not even solved by methods that fall into the
category of artificial intelligence. Hand-crafted behavior is the state
of the art for about two decades. \citet{Roefer2018} state that
``In the domain of RoboCup, real-time requirements and limited computational
resources often prevent the use of planning-based approaches''.
Between 2009 and 2017 three distinct teams won the RoboCup Standard
Platform League (SPL), which is carried out every year. All of them used
rather static behaviors: B-Human, UT Austin Villa, and rUNSWift.
Few information about the behaviors used by UT Austin Villa is available but
the report accompanying their code release \citep{Barrett2012} suggests that
behavior is hand-crafted.
rUNSWift's behavior is hand-crafted and written in Python \citep{Bianchi2015}.
B-Human used XABSL \citep{Loetzsch2006} and currently uses CABSL
\citep{Roefer2018} to describe behaviors. Both languages are used to define
hierarchical finite state machines for the robots' behavior.
Only in 2018 a team using a ``dynamic strategy'', Nao-Team HTWK, won the
RoboCup SPL. They represent the problem of positioning players that are not
close to the ball as an optimization problem and solve it \citep{Mewes2014}.
That, however, is only a part of the whole soccer behavior.

\subsection{Complexity of Systems Is Increasing}

Over the years, complexity of robotic systems and the posed problems increased.
A complex six-legged walking robot had 12 DOF \citep{Maes1990} at the
beginning of the 90s. In 2016, a quadrupedal robot with two arms for
manipulation had to handle 61 DOF \citep{Bartsch2016}. Controlling such
a complex robot is still a challenging problem. Most of the
presented works in the field of manipulation only have to handle
six or seven DOF while complex robots control 17 \citep{Kormushev2011b}
or 24 DOF \citep{Bartsch2016} to generate a walking behavior
or 24 DOF for in-hand manipulation \citep{Rajeswaran2018,Andrychowicz2018}.
For comparison, a well-studied biological system is the human body.
It has an estimated total number of 244 DOF and a
conservatively estimated number of 630 skeletal muscles \citep{Zatsiorsky2012}.
It is, hence, a much more complex system to control than any of the
robots that have been used in the works that we refer to in this survey.
There is still a long way to go to reach the same level of flexibility
and agility.

Not only the actuation capabilities are improving but also the complexity of
used sensors increased considerably in almost three decades of behavior
learning research on real robots. In early applications only very simple
sensors have been used, for example, four light sensors \citep{Kirchner1997}.
Alternatively, the perception problem has been decoupled from the action problem to solve
it with computer vision and state estimation \citep{Muelling2013,Parisi2015}.
In more recent works, raw camera images have been used directly by the
learned behavior \citep{Lampe2013,Levine2016,Levine2017} and RGB-D cameras
have been used \citep{Lenz2015}.
RGB-D cameras are probably the most complex sensors that are used in learned
behaviors today.
Robotics research in general is already more advanced and we will see
other complex sensors in addition to rather conventional cameras.
For example, current robotic systems can have advanced tactile sensor arrays
based on fiber-optic sensing principles \citep{Bartsch2016}.

\subsection{Limited Versatility of Learned Skills}

The works on bipedal walking are particularly interesting, since they allow
a direct comparison of the application on real robots and the application
in simulation and computer graphics.
\citet{Peng2017} learned bipedal walking on two levels: a low-level walking
behavior and a high-level behavior that generates the walking direction.
The high-level behavior incorporates information about the surrounding
terrain and has been used to follow trails, dribble a soccer ball towards a
target, and navigate through static and dynamic obstacles. The low-level
behavior only knows about the internal state of the walker and the desired
goal of the high-level behavior and was trained to be robust against
disturbances and terrain variations.
Also \citet{Peng2018} demonstrate how imitation and reinforcement learning
can be used to generate realistic acrobatic movements: performing a cartwheel,
backflip, frontflip, roll, vault, dancing, kicking, punching, standing up, etc.
Those skills are then combined to a complex sequence of behaviors.
In comparison, learned biped walking behaviors on real robots are usually only
tested in controlled environments in the lab \citep{Benbrahim1997,Matsubara2005,
Geng2006,Kormushev2011b,Missura2015}.


Walking is just one example of how skills that have been learned on real
robots are often not versatile.
Another example is grasping: the currently
most impressive work, published by \citet{Levine2017}, is applicable to
a large variety of objects but only if the camera is in a certain angle
to the objects and only vertical pinch grasps have been considered.
Other behaviors, for example, tee-ball \citep{Peters2005,Peters2008},
pancake flipping \citep{Kormushev2010b}, plugging in a power plug
\citep{Chebotar2017}, flipping a light switch \citep{Buchli2011},
do not even include the position of the manipulated object in their
control loop. Many of the learned behaviors are hence still only applicable
under controlled lab conditions.

\subsection{Limited Variety of Considered Problems}

In natural learning agents (also known as animals), there is evidence that
the same learning mechanisms can be evolved and used to solve a variety of
tasks:
``A major role of the early vertebrate CNS [central nervous system] involved
the guidance of swimming based on receptors that accumulated information from
a relatively long distance, mainly those for vision and olfaction. The original
vertebrate motor system later adapted into the one that controls your reaching
and pointing movements.'' \citep[page 9]{Shadmehr2005}

In behavior learning for robots, however, often the same simple problems are
tackled again and again with only minor variations but with a large variety
of different learning algorithms.
Learning efforts often focus on grasping, walking, and batting.
Certainly, these problems are not solved yet
(\citet{Johns2016}: ``Robot grasping is far from a solved problem.'').
Furthermore, solving the exact same problem again is good for benchmarking.
Yet, the variety of problems solved by learning is low.
We should also try to solve a larger variety of problems to discover and
tackle new challenges in behavior learning and to improve our set of tools.
Examples are given in the outlook.

Most of the considered problems are also low-level motor skills. While this
seems to be too simple at first, there is also a justification for it.
\citet[page 1]{Shadmehr2005} state that motor learning,
that is, learning of low-level behavior, uses the same basic mechanisms as
higher forms of intelligence, for example, language and abstract reasoning.
However, the goal should be to demonstrate that learning is possible and
useful at all levels of behavior and to actually use its full potential.

\subsection{Reasons for Current Limitations}

What hinders robots from learning the same skills as humans with a similar
performance these days? There are several reasons.
We identify the main reasons as algorithmic, computational, and hardware
problems.

One of the most advanced fields of artificial intelligence is computer
vision based on deep learning. In some specific benchmarks, computer
vision is better than humans but it is not as robust as a human which
has been demonstrated with adversarial examples \citep{Szegedy2013}.
In addition, semantic segmentation, tracking objects in videos,
object detection with a large amount of classes are examples for very
active research topics in which humans are a lot better. Computer vision
is one example of a domain which behavior learning builds upon. When
we learn grasping \citep{Levine2017} or visual servoing \citep{Levine2016}
end-to-end, we make use of the results from computer vision research.
While we do not reach human-level performance in these areas, we can hardly
surpass it in real-world behavior learning problems.
Also reinforcement learning algorithms are not yet at the point where
they are sample-efficient enough to learn complex behaviors from a reasonable
amount of data. One of the most impressive works in this field
at the moment is from \citet{Andrychowicz2018}. They learned complex
in-hand manipulation skills to rotate a cube into any desired orientation.
Approximately 100 years of experience were used during the training process.
Still the robustness of the skill is not comparable to an average human:
on average 26.4 consecutive rotations succeed when 50 is the maximum length
of an experiment. Certainly no human spent 100 years on learning exclusively
in-hand manipulation to reach a much better level of performance.

Many state-of-the-art algorithms in machine learning have also high
demands on processing power during prediction phase
\citep{Silver2016,Levine2017,Andrychowicz2018}
which makes them slow in reaction time, maybe not even suitable for
autonomous systems that have to budget with energy, and training on
a robotic system might be infeasible.

Probably the main reason why so many researchers do not learn complex skills
for robots in reality is that robots break too easily.
Absence of training data from dangerous situations is a problem.
It motivated \citet{Gandhi2017} to record a datasets of drones crashing
into obstacles.
In contrast, humans fail and fall all the time and gain lots of negative
experiences. There is probably not a single
professional soccer match that has been played over the full length in which no
player is falling down unexpectedly and, yet, most players are not seriously
injured.
Humans are colliding all the time with objects when they move things
around, for example, while eating at an overly full dinner table.
The difference is that humans are flexible, soft, and lightweight.
A human is lightweight compared to similarly strong robots. Humans' force to
weight ratio is much better. The best Olympic weight lifters can move weights
that are more than twice as heavy as they are. Humans are extremely flexible.
As already mentioned, they have about 244 DOF and 630 skeletal muscles
\citep{Zatsiorsky2012} and most of their body is soft
while one of the most complex robots today has 61 DOF and
consists mostly of stiff and rigid parts \citep{Bartsch2016} that are at
the same time very fragile.
A new actuation paradigm is required for robots that solve dynamic, partially
observable problems. \citet{Haddadin2009} propose to use elastic joints in
the domain of robot soccer. Elastic joints make robots more robust,
collaboration or competition with humans safer, and they would enable higher
maximum joint speeds. Controlling elastic joints is more complex though.
In addition, humans have many sensors (tactile, acoustic, vestibular) that are
used to recognize unexpected events and they can react accordingly:
they learned to fall or to stop moving the arm before they pull down
the bottle from the dining table.

\section{Outlook}
\label{s:outlook}

We will conclude with several advices that we find are important and
an outlook on future behavior learning problems that could be tackled.

\subsection{Ways to Simplify Learning Problems}

\citet{Kirchner1997} states: ``we believe that learning has to be used but it
needs to be biased. If we attempt to solve highly complex problems, like the
general robot learning problem, we must refrain from tabula rasa learning and
begin to incorporate bias that will simplifies [sic] the learning process.''

Ways to simplify the learning problem are to not learn everything from scratch
(knowledge transfer), not everything end-to-end (combination with other methods),
to learn while a safe, deliberative method is operating, or to learn in a
controlled environment (bootstrapping).

\paragraph{Knowledge transfer:}
Knowledge can be transferred from similar tasks, similar systems, or similar
environments. In the optimal case multiple almost identical robots are used
to learn the same task in the same environment \citep{Gu2016,Levine2017}.
\citet{Levine2017} also show that data transfer from one robot to another
robot in the same environment, solving a similar task, is beneficial
(if actions are represented in task space). \citet{Levine2016} also show
that pretraining is a key factor for success when very complex behaviors
are trained end-to-end.

In our opinion, more research should be done on lifelong learning.
It could lead to robust, sample-efficient artificial intelligence that
is able to solve a multitude of tasks and, hence, share knowledge.
Lifelong learning is defined by \citet{Silver2013}:
``Lifelong Machine Learning, or LML, considers systems that can learn many
tasks over a lifetime from one or more domains. They efficiently and
effectively retain the knowledge they have learned and use that knowledge to
more efficiently and effectively learn new tasks.''
We believe that this can be much more efficient than learning everything
from scratch.
Coming back to the example of in-hand manipulation \citep{Andrychowicz2018},
perceiving the object's pose or several strategies used in the manipulation
behavior are components that could be shared with many other tasks that
are related to manipulation of movable objects.

We have to find ways to share knowledge between similar and dissimilar robots
and between similar and dissimilar tasks. In theory, sharing knowledge between
robots in form of training sets or pretrained models is much easier than
sharing knowledge between humans that can only absorb knowledge
through their senses.
\citet{Bozcuoglu2018} propose a similar approach: they share ontologies
and execution logs on the cloud platform openEASE. The knowledge can be
transferred to other environments or other robots.
The same approach could be used to share pretrained models or training
data to learn behaviors.

\paragraph{Combination with other methods:}
Combining existing approaches for perception and state estimation with
machine learning has been shown to be effective by
\citet{Muelling2013,Parisi2015}. Similarly, combining existing approaches for planning
and machine learning has been shown to be effective by \citet{Lenz2015}.
Also model predictive control has been combined with a learned
uncertainty-aware perception model by \citet{Kahn2017}.
\citet{Nemec2017} combine machine learning and structured search with
physical constraints.
To generate walking behaviors, often classical models like a linear inverted
pendulum \citep{Kajita2001} are used, a zero moment point
\citep{Vukobratovic2005} is computed.
Mostly, only parts of complex walking behaviors are learned.
We think this is still a valid method to verify and understand what is
happening on the system, to reduce the amount of physical interaction
with the world that is required to learn the behavior and to obtain solutions
that are more safe.
\citet{Geng2006} confirm this for their application. They state that:
``Building and controlling fast biped robots demands a deeper understanding
of biped walking than for slow robots.''
\citet{Englert2018} state:
``One way to reduce ... difficulties is by exploiting the problem structure
and by putting prior knowledge into the learning process.''
Although \citet{Loquercio2018} show remarkable results of an almost
end-to-end learning approach for collision avoidance on a drone.
They do not want to replace ``map-localize-plan'' approaches
and believe that ``learning-based and traditional approaches will one day
complement each other''.
An example of a promising idea that shows how established methods can be
combined with machine learning is the incorporation of Kalman filters
in a neural network. This approach has been presented by \citet{Kassahun2008}.

However, we have to make sure that we do not artificially limit the
amount of learnable behaviors by introducing too strong constraints or too
simple models.
For example, requiring the zero moment point (ZMP) to be in a support
polygon is a strong restriction. It is an artificially constructed,
simple model of dynamical stability,
that is developed to avoid at all costs that expensive robots fall and break.
It limits the capabilities of a
robot, for example, running would be very hard to implement with a ZMP
approach.
Furthermore, \citet{Yang2017} state that this approach prohibits advanced
balancing behaviors.
Making basic physical knowledge available to the learning algorithm
can be beneficial without restricting the amount of learnable behaviors
though.
As an alternative to the ZMP approach, we can compute the centroidal momentum
\citep{Orin2008,Orin2013} and make it available to the learning algorithm.
When a translation from joint space to Cartesian space is required or useful,
we can use the Jacobian.
For dynamics we can make use of the equations of motion.

\paragraph{Boostrapping:}
An obvious situation where the combination of behavior learning with another
method is safer is manipulation with a superimposed collision avoidance
behavior. While the robot is learning to grasp, it can safely be guided
around obstacles. These ``safety mechanisms'' could also be used to bootstrap
learning and collect data safely before we shift to the pure learned behavior
that might perform better. It is even possible to use additional equipment or a
controlled environment to provide additional information to bootstrap
learning. This has been done, for example, by \citet{Levine2016} to reduce
the required amount of data. \citet{Englert2018} also demonstrate that
a combination of optimal control, episodic reinforcement learning, and
inverse optimal control in the training phase can be safe and efficient.
The problem of safe exploration has also been discussed in more detail
by \citet[pages 14--17]{Amodei2016}.

\subsection{Comparability and Reproducibility}


\citet{Shadmehr2005} convey the idea that the same computational principles
that allow earlier forms of life to move in their environment later enabled
higher forms of intelligence like language and reasoning.
The intelligence of animals and humans evolves with the complexity of the
problems that it solves. An example for this is confirmed by
\citet{Faisal2010}: the production of
early prehistoric (Oldowan) and later (Acheulean) stone tools
has been investigated. Oldowan tools are simpler and their production require
less complex behaviors. The production of Acheulean tools requires
the activation of brain regions associated with higher-level behavior
organization. The development of more complex behavior coordination
could even be linked to the development of more complex forms of
communication.
The development of complex manipulation behaviors required more intellectual
capacities. These could also be applied to another domain -- in this case:
language.
This is an important finding for us as roboticists.
Translating this to our work, this means more complex problems require the
development of better behavior learning algorithms.
These algorithms could potentially also be used in other domains for which
they have not been directly designed.
Hence, advancing at both frontiers could benefit the whole field.

Artificial intelligence has advanced by setting challenging benchmark
problems. For example, the problem of playing chess against a human
or the RoboCup initiative that has a similar goal but combines AI with
robotics \citep{Kitano1997}:
``The Robot World-Cup Soccer (RoboCup) is an attempt to foster AI and
intelligent robotics research by providing a standard problem where a wide
range of technologies can be integrated and examined.''
In recent years we have seen major advances in reinforcement learning
also because clearly defined benchmarks are available, for example,
the Atari learning environment \citep{Bellemare2012} and OpenAI Gym
\citep{Brockman2016}.
These benchmarks make comparisons of existing approaches easier.
It is also simpler to reproduce results because it is easy to check
if a reimplementation of an algorithm gives the same result as in
the original publication.
Hence, we recommend to define benchmarks for robotic behavior learning.

A problem is that often similar problems are solved but with
varying conditions, for example, in the context of grasping we observed
that the objects are often different although there are standardization
efforts: the YCB object and model set is an example
\citep{Calli2015a,Calli2015b,Calli2017}.
These efforts have to be fostered and supported. Also new benchmarks
have to be created. We can also learn here from the diagnosis and treatment
of human patients. An example for a ``benchmark'' for humans is the
box and block test \citep{Mathiowetz1985}: the patient has to move colored
blocks from one box to another as fast as possible.
We think that a set of benchmark problems should be selected, standardized,
formalized, and described in detail so that results are easily comparable.


Games and sports are particularly good candidates for benchmark problems
because they have a clear set of rules, standardized material, they are
usually easy to understand, and offer a variety of challenging problems.
We have seen that a large number of behavior learning problems
already come from this domain. Mostly subproblems like kicking or
batting a ball have been extracted and learned.
More advanced benchmarks would also include tasks with less strict rules,
for example, setting a table.

Benchmarking in the context of robotics, however, is difficult because software
can usually not be tested in isolation. Simulations could be used to address
this problem but they often lead to solutions that are not transferable
to reality, neither the learned behavior nor the learning algorithm.
The RoboCup Standard Platform League (SPL) solves this problem by requiring
that each competing team uses the same hardware. This is not an optimal
solution because most robots are expensive and research institutes are usually
not able to buy a new robot just to compete in a specific benchmark.
We can offer no perfect solution for this problem.
We can only propose that a cheap robotic platform that is
sufficient enough for a variety of benchmarks should be developed.

\subsection{The Future of Behavior Learning Problems}

\citet{Mason2012} writes:
``What percentage of human's manipulative repertoire have robots mastered?
Nobody can answer this question.''
We can say exactly the same about any other category of robotic behaviors.
At least we now have a rough overview of which behaviors have been learned.
We will now try to talk about what is still missing.


At the moment, most behaviors are learned in isolation. On a complete system,
the learned behavior will interfere with high-level behaviors and other
behaviors on the same level that might even have higher priority, for example,
balancing or collision avoidance.
There might even be other learned behaviors,
for example, a learned walking behavior and a learned throwing behavior could
be executed in parallel. Executing multiple behaviors in parallel has effects
on the whole system. These problems are neglected if behaviors are learned
in isolation. Throwing a ball while walking makes the balancing part of the
walking behavior more difficult and grasping an object while collision
avoidance is active might result in different reaching trajectories.
Sometimes combining two behaviors might require one of these behaviors
to be changed completely. For example, in the case of throwing while
running, the whole locomotion and balancing behavior might have to be
altered to absorb high forces that are exerted during the throw.

\begin{figure*}[tb]
\centering
\includegraphics[width=\textwidth]{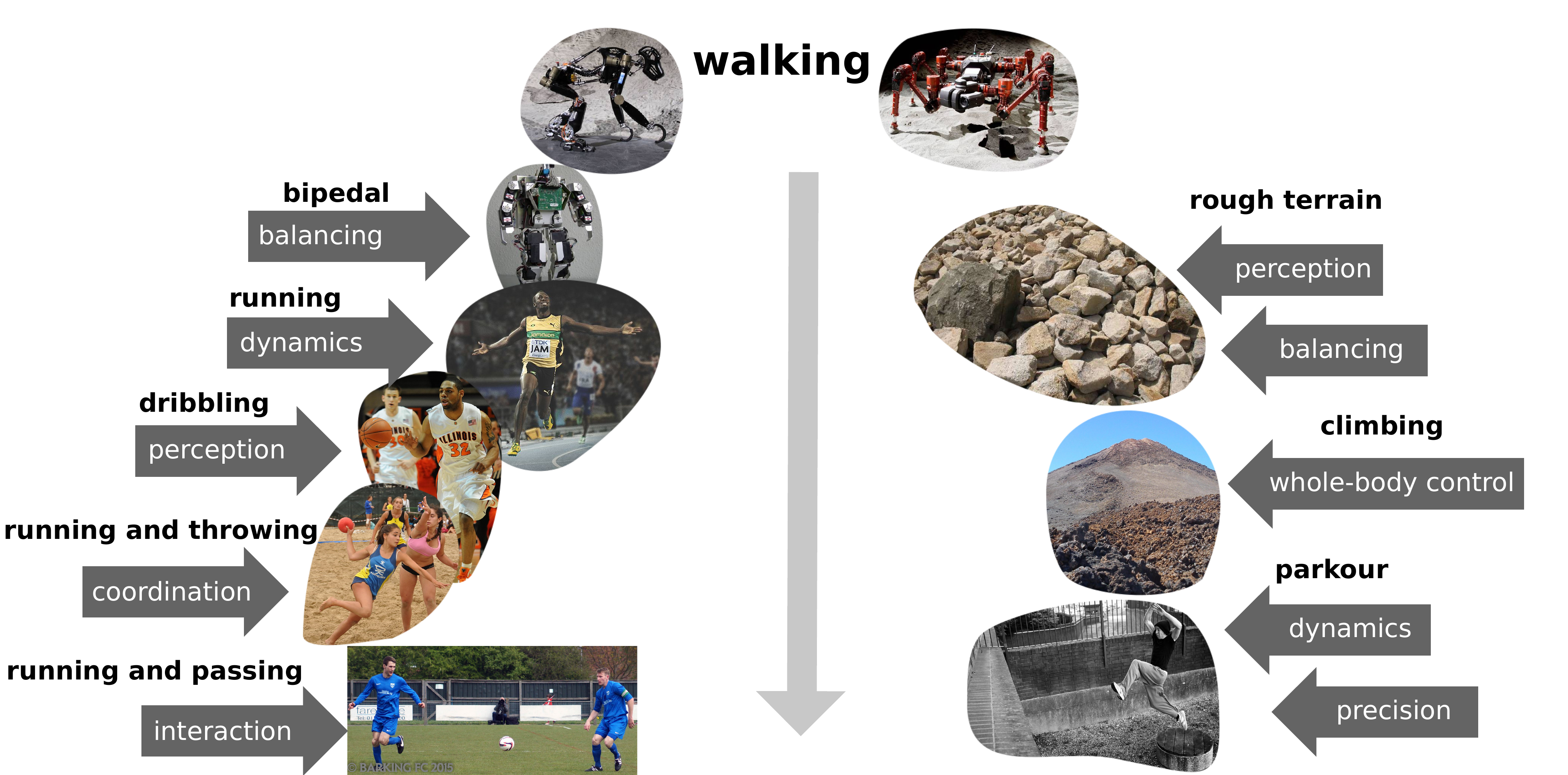}
\caption{\textbf{Roadmaps for walking robots.}
{
Sources:
running from Stephane Kempinaire (URL:\linebreak \url{http://www.mynewsdesk.com/se/puma-nordic/images/puma-aw14_ff_bolt-325510}; license: CC BY 3.0),
dribbling from flickr user tsavoja (URL: \url{https://www.flickr.com/photos/tsavoja/4106568938/}; license: CC BY-SA 2.0),
throwing while running from flickr user RFEBM Balonmano (URL: \url{https://www.flickr.com/photos/125948220@N02/14826033503/}; license: CC BY-SA 2.0),
passing while running from flickr user Terry Gilbert (URL: \url{https://flic.kr/p/QDhaKN}; license: CC BY 2.0),
parkour from flickr user THOR (URL:\linebreak \url{https://www.flickr.com/photos/geishaboy500/3090363361/}; license: CC BY 2.0),
all other photos are from DFKI RIC and can be found at \url{https://robotik.dfki-bremen.de/}
}
\label{fig:outlook_walking}
}
\end{figure*}

Figure \ref{fig:outlook_walking} illustrates two possible roadmaps for walking
behaviors. Currently, we are able to learn walking with quadrupedal or
six-legged robots. There a two alternative routes illustrated that we could
take from there: the ``ball sports route'' and the ``parkour route''.
Ball sports in this example include soccer, basketball, or handball.
It is to some extent possible to learn bipedal walking, which requires
more advanced balancing behavior than walking with more legs. Fast bipedal
running is already a much more complex task because it is a highly dynamic
behavior that cannot easily be solved with classical stability criteria and
control approaches. Running and dribbling a ball requires to solve a much
more complex perception problem and precise foot placement or hand movements.
Combining this behavior with the requirement to throw or kick a ball will
introduce a difficult coordination problem: throwing will have an impact on the
balancing part of the running behavior. A good solution will predict
this impact and counteract already while the throw is performed.
However, throwing a ball to a fixed goal is easy in comparison to passing
the ball to a teammate. In this case, the robot has to anticipate the
behavior of the teammate to pass the ball to a location where the teammate
will be able to make use of it.
Another future research direction could be over climbing to parkour. Legged
robots unfold their full potential in rough and irregular terrain, where
precise perception of the environment, foot placement and robust balancing is
required. This has been learned already to some extent. A more difficult
scenario would be climbing up a mountain with steep slopes, where not only
feet but all body parts must be controlled, for example, a humanoid would
have to use its arms. The robot must be flexible enough to balance on
steep and rough terrain. A next possible step would be one of the most
difficult sports that humans are able to perform: parkour. It requires to
``understand'' the environment, that is, know what you can do with it to find
the fastest and direct way by overcoming obstacles. The whole body is involved
and it is often required to turn off basic safety mechanisms, for example,
to perform a double kong vault where the body is almost turned upside down
with the hands on the obstacle directing momentum and the feet above the
head to get out of the way.

There are low-hanging fruits to increase the spectrum of learned behaviors.
Examples are the locomotion behaviors running, climbing ladders, jumping over
obstacles, jumping precisely or jumping as high as possible with one or two
legs, front or back flip, swimming, and paddling.
In the kitchen domain stirring, chopping, opening cans or bottles.
In the household domain the problems of folding sheets or clothes can be
very challenging because these problems are very hard to model.
In the manufacturing domain the skills of hammering, sawing, sewing,
splitting wood, shoveling, drilling, and tool use in general are relevant.
While perception has been fully learned, for example, for grasping and
collision avoidance, this has not been considered so far for very dynamic
problems like catching balls, batting or kicking balls, etc. There is a
limited amount of publication concerned with learning high-level game
playing in real physical games, for example, to learn coordination of
multiple robots in soccer.
For interaction with humans, performing gestures and other physical interaction
behaviors, for example, various forms of hand shaking could be learned.
Interesting balancing problems often come from sports like surfing, skating,
or skiing.

There are not many learned behaviors that require advanced spatio-temporal
and causal reasoning beyond unscrewing a light bulb. Assembling furniture,
tidying up a room, cooking a complete meal, or solving puzzles are examples
for these kind of problems.

Creating a system that solves not just one problem but a variety of
complex tasks is even more difficult.
It involves integration of hardware components, software components,
and behaviors. Building complex systems is a challenge in itself,
but it is required to create more sophisticated complex behaviors.

Learned behaviors can usually not be explained. Robots cannot reason
about them. They cannot explain why they selected a certain action
or why it works.
We have not yet seen robots that combine existing learned behaviors
to new sequences or combinations of behaviors to solve tasks that they
have not seen before.

Given the current development in behavior learning and in computer
vision, we expect that the next big steps will be made by deep learning
and by solving more and more complex perception problems.
This direction of artificial intelligence research has its justification in
Moravec's paradox:
``it is comparatively easy to make computers exhibit adult level performance
on intelligence tests or playing checkers, and difficult or impossible to give
them the skills of a one-year-old when it comes to perception and mobility''
\citep[p. 15]{Moravec1988}.
However, we emphasize that for complex behaviors not only
complex perception but also complex control is required. It is not
sufficient to control a 7 DOF arm to realize a versatile, flexible,
and autonomous humanoid robot.
We should strive towards pushing the limits in terms of kinematic complexity
like the work of \citet{Andrychowicz2018}, who control a complex, human-like
hand.

In summary, there is still a long way to go to build robots that are able to
perform as good as humans in these tasks but we think that learning behaviors
is one of the best ways that we have to acquire these skills when the
robotic hardware is sufficient enough.
\citet{Mason2012} formulated a conjecture about robotics research:
``[..] it is just possible that our field is still in its infancy.
I do not have a compelling argument for this view, but it is telling that
we have no effective way to measure our progress toward long-range goals.''
Our outlook on which skills we should try to master by behavior learning
in the future, particularly the discussion of the roadmap displayed in
Figure \ref{fig:outlook_walking}, also is a confirmation of this.

\section*{Acknowledgments}
This work was supported through grants
from the European Union's Horizon 2020 research and innovation program (No H2020-FOF 2016 723853),
from the German Federal Ministry for Economic Affairs and Energy (BMWi, No 50RA1703),
and from the German Federal Ministry of Education and Research (BMBF, No 01IW18003).

We thank Hendrik Wiese, Elsa Andrea Kirchner, Matias Valdenegro-Toro and
Malte Wirkus for useful hints and discussions.
The definitions of skill and motion plan that we use in this article have been
developed in discussions with Elsa Andrea Kirchner, Lisa Gutzeit,
Jos{\'e} de Gea Fern{\'a}ndez, Alexander Dettmann, Sebastian Stock,
Dennis Mronga, Nils Niemann and Sebastian Bartsch
whom we would like to thank for their contributions.
We would like to thank particularly Jos{\'e} de Gea Fern{\'a}ndez and
Thomas M. Roehr for their valuable feedback.

\newcommand{\icp}{&{\textperiodcentered}&{\textperiodcentered}}

\newpage \clearpage
\begingroup
\small\sf\centering
\begin{longtable}{ll|ll|ll|ll|ll|ll|ll}
\caption{
\textbf{Overview of learned behaviors.}
\label{tab:complete}}\\
\toprule
\bfseries Behavior & \bfseries Publication &
\bfseries \rotatebox{90}{Perception \dag} &
\bfseries \rotatebox{90}{Action \dag} &
\bfseries \rotatebox{90}{Deliberative \ddag} &
\bfseries \rotatebox{90}{Reactive \ddag} &
\bfseries \rotatebox{90}{Discrete} &
\bfseries \rotatebox{90}{Rhythmic} &
\bfseries \rotatebox{90}{Static} &
\bfseries \rotatebox{90}{Dynamic} &
\bfseries \rotatebox{90}{Active} &
\bfseries \rotatebox{90}{Passive} &
\bfseries \rotatebox{90}{Locomotion} &
\bfseries \rotatebox{90}{Manipulation}\\
\midrule
\endfirsthead
\caption{\textbf{Overview of learned behaviors} \textit{(continued)}}\\
\toprule
\bfseries Behavior & \bfseries Publication &
\bfseries \rotatebox{90}{Perception \dag} &
\bfseries \rotatebox{90}{Action \dag} &
\bfseries \rotatebox{90}{Deliberative \ddag} &
\bfseries \rotatebox{90}{Reactive \ddag} &
\bfseries \rotatebox{90}{Discrete} &
\bfseries \rotatebox{90}{Rhythmic} &
\bfseries \rotatebox{90}{Static} &
\bfseries \rotatebox{90}{Dynamic} &
\bfseries \rotatebox{90}{Active} &
\bfseries \rotatebox{90}{Passive} &
\bfseries \rotatebox{90}{Locomotion} &
\bfseries \rotatebox{90}{Manipulation}\\
\midrule
\endhead
flipping a light switch & \citet{Buchli2011} & \xmark & \cmark & \cmark & \xmark & \cmark & \xmark & \cmark & \xmark & \cmark & \xmark & \xmark & \cmark \\
\midrule
open door & & & & & & \cmark & \xmark & \cmark & \xmark & \cmark & \xmark & \xmark & \cmark\\
& \citet{Kalakrishnan2011} & \cmark & \cmark & \xmark & \cmark \icp\icp\icp\icp \\
& \citet{Gu2016} & \cmark & \cmark & \xmark & \cmark \icp\icp\icp\icp \\
& \citet{Kormushev2010,Kormushev2011} & \cmark & \cmark & \xmark & \cmark \icp\icp\icp\icp \\
& \citet{Nemec2017} & \xmark & \cmark & \cmark & \xmark \icp\icp\icp\icp \\
& \citet{Chebotar2017b} & \cmark & \cmark & \xmark & \cmark \icp\icp\icp\icp \\
& \citet{Englert2018} & \xmark & \cmark & \xmark & \xmark \icp\icp\icp\icp \\
\midrule
valve turning & \citet{Carrera2012} & \xmark & \cmark & \cmark & \xmark & \cmark & \xmark & \cmark & \xmark & \cmark & \xmark & \xmark & \cmark \\
crank-turning & \citet{Petric2014} & \cmark & \cmark & \xmark & \cmark & \xmark & \cmark & \cmark & \xmark & \cmark & \xmark & \xmark & \cmark \\
screw cap on bottle & \citet{Levine2016} & \cmark & \cmark & \xmark & \cmark & \cmark & \xmark & \cmark & \xmark & \cmark & \xmark & \xmark & \cmark \\
\midrule
peg-in-a-hole &&&&&& \cmark & \xmark & \cmark & \xmark & \cmark & \xmark & \xmark & \cmark \\
& \citet{Gullapalli1994} & \cmark & \cmark & \xmark & \cmark \icp\icp\icp\icp \\
& \citet{Ellekilde2012} & \xmark & \cmark & \cmark & \xmark \icp\icp\icp\icp \\
& \citet{Levine2016} & \cmark & \cmark & \xmark & \cmark \icp\icp\icp\icp \\
& \citet{Kramberger2016} & \cmark & \cmark & \cmark & \xmark \icp\icp\icp\icp \\ 
$\vdash$ power plug & \citet{Chebotar2017} & \xmark & \cmark & \xmark & \xmark \icp\icp\icp\icp \\
$\vdash$ connect a pipe & \citet{Laursen2018} & \xmark & \cmark & \cmark & \xmark \icp\icp\icp\icp \\
\midrule
ironing & \citet{Kormushev2010,Kormushev2011} & \cmark & \cmark & \xmark & \cmark & \cmark & \xmark & \cmark & \xmark & \cmark & \xmark & \xmark & \cmark\\ 
whiteboard cleaning & \citet{Kormushev2011c} & \cmark & \cmark & \xmark & \cmark & \cmark & \xmark & \cmark & \xmark & \cmark & \xmark & \xmark & \cmark \\ 
grinding / polishing & \citet{Nemec2018} & \cmark & \cmark & \xmark & \cmark & \cmark & \xmark & \cmark & \xmark & \cmark & \xmark & \xmark & \cmark\\ 
wiping & &&&&&&&&&&&& \\
& \citet{Urbanek2004} & \xmark & \cmark & \xmark & \xmark & \xmark & \cmark & \cmark & \xmark & \cmark & \xmark & \xmark & \cmark \\
& \citet{Gams2014} & \xmark & \cmark & \xmark & \cmark & \xmark & \cmark & \cmark & \xmark & \cmark & \xmark & \xmark & \cmark \\
sweeping &&&&&&&&&&&&&\\
& \citet{Alizadeh2014} & \xmark & \cmark & \cmark & \xmark & \cmark & \xmark & \xmark & \cmark & \cmark & \xmark & \xmark & \cmark \\
& \citet{Pervez2017} & \cmark & \xmark & \xmark & \cmark & \cmark & \xmark & \xmark & \cmark & \cmark & \xmark & \xmark & \cmark \\
\midrule
handwriting &&&&&&&&  &  & \cmark & \xmark & \xmark & \cmark \\
& \citet{Manschitz2018} & \xmark & \cmark & \xmark & \xmark & \cmark & \xmark & \cmark & \xmark \icp\icp \\
& \citet{Berio2016} & \xmark & \cmark & \xmark & \xmark & \cmark & \xmark & \xmark & \cmark \icp\icp \\
\midrule
calligraphy & \citet{Omair2015} & \cmark & \cmark & \xmark & \xmark & \cmark & \xmark & \xmark & \cmark & \cmark & \xmark & \xmark & \cmark \\
\midrule
grasping &&&&&& \cmark & \xmark & \cmark & \xmark & \cmark & \xmark & \xmark & \cmark \\
& \citet{Steil2004} &
\xmark & \cmark & \xmark & \xmark
\icp\icp\icp\icp  \\
& \citet{Kroemer2009} &
\xmark & \cmark & \xmark & \cmark 
\icp\icp\icp\icp  \\
& \citet{Grave2010} &
\xmark & \cmark & \xmark & \xmark
\icp\icp\icp\icp  \\
& \citet{Stulp2011} &
\xmark & \cmark & \xmark & \xmark
\icp\icp\icp\icp  \\
& \citet{Kalakrishnan2011} &
\cmark & \cmark & \xmark & \cmark \icp\icp\icp\icp  \\
& \citet{Amor2012} &
\xmark & \cmark & \cmark & \xmark
\icp\icp\icp\icp  \\
& \citet{Lampe2013} &
\cmark & \cmark & \xmark & \cmark \icp\icp\icp\icp  \\
& \citet{Lenz2015} &
\cmark & \xmark & \cmark & \xmark \icp\icp\icp\icp  \\
& \citet{Pinto2016} &
\cmark & \xmark & \cmark & \xmark \icp\icp\icp\icp  \\
& \citet{Johns2016} &
\cmark & \xmark & \cmark & \xmark \icp\icp\icp\icp  \\
& \citet{Levine2017} &
\cmark & \cmark & \xmark & \cmark \icp\icp\icp\icp  \\
& \citet{Mahler2017} &
\cmark & \cmark & \xmark & \cmark \icp\icp\icp\icp  \\
\midrule
pick \& place &&&&&& \cmark & \xmark & \cmark & \xmark & \cmark & \xmark & \xmark & \cmark \\
& \citet{Stulp2012} & \xmark & \cmark & \xmark & \xmark \icp\icp\icp\icp \\
& \citet{Ijspeert2013} & \xmark & \cmark & \cmark & \xmark \icp\icp\icp\icp \\
& \citet{Rahmatizadeh2016} & \xmark & \cmark & \cmark & \xmark
\icp\icp\icp\icp \\
& \citet{Chebotar2017b} & \cmark & \cmark & \xmark & \cmark \icp\icp\icp\icp \\
& \citet{Kroemer2017} & \xmark & \cmark & \cmark & \xmark  
\icp\icp\icp\icp \\
& \citet{Levine2016} & \cmark & \cmark & \xmark & \cmark \icp\icp\icp\icp \\
& \citet{Finn2017} & \cmark & \cmark & \xmark & \cmark \icp\icp\icp\icp \\
\midrule
block stacking &&&&&& \cmark & \xmark & \cmark & \xmark & \cmark & \xmark & \xmark & \cmark \\
& \citet{Deisenroth2015} & \xmark & \cmark & \cmark & \xmark
\icp\icp\icp\icp \\
& \citet{Duan2017} & \cmark & \cmark & \xmark & \cmark
\icp\icp\icp\icp \\
\midrule
in-hand manipulation &&&&&&&&&& \cmark & \xmark & \xmark & \cmark \\
& \citet{Hoof2015} & \xmark & \cmark & \xmark & \cmark & \cmark & \xmark & \cmark & \xmark \icp\icp \\
& \citet{Rajeswaran2018} & \cmark & \cmark & \xmark & \cmark & \cmark & \xmark & \cmark & \xmark \icp\icp \\
& \citet{Andrychowicz2018} & \cmark & \cmark & \cmark & \xmark & \cmark & \xmark & \cmark & \xmark \icp\icp \\
\midrule
tumbling / tilting objects &&&&&&&&&&&&&\\
& \citet{Pollard2004} & \xmark & \cmark & \cmark & \xmark & \cmark & \xmark & \cmark & \xmark & \cmark & \xmark & \xmark & \cmark \\
& \citet{Kroemer2017} & \xmark & \cmark & \xmark & \xmark & \cmark & \xmark & \cmark & \xmark & \cmark & \xmark & \xmark & \cmark \\
\midrule
hockey & & & & & & \cmark & \xmark & \xmark & \cmark & \cmark & \xmark & \xmark & \cmark \\
& \citet{Daniel2013} & \xmark & \cmark & \xmark & \xmark
\icp\icp\icp\icp \\
& \citet{Chebotar2017} & \xmark & \cmark & \xmark & \cmark \icp\icp\icp\icp \\
& \citet{Rakicevic2017} & \xmark & \cmark & \xmark & \xmark
\icp\icp\icp\icp \\
& \citet{Paraschos2018} & \xmark & \cmark & \xmark & \xmark
\icp\icp\icp\icp \\
\midrule
knot tying & \citet{vandenBerg2010} & \xmark & \cmark & \xmark & \xmark & \cmark & \xmark & \cmark & \xmark & \cmark & \xmark & \xmark & \cmark\\
knot untying & \citet{wenHaoLui2013}
 & \cmark 
 & \xmark 
 & \xmark 
 & \xmark 
 & \cmark 
 & \xmark 
 & \cmark 
 & \xmark 
 & \cmark 
 & \xmark 
 & \xmark 
 & \cmark\\ 
\midrule
folding a shirt & \citet{Colome2018} & \xmark & \cmark & \xmark & \xmark & \cmark & \xmark & \cmark & \xmark & \cmark & \xmark & \xmark & \cmark \\
holding garment & \citet{Corona2018} & \cmark & \xmark & \cmark & \xmark & \cmark & \xmark & \cmark & \xmark & \cmark & \xmark & \xmark & \cmark \\
dressing assistance & \citet{Erickson2018} & \xmark & \xmark & \cmark & \xmark & \cmark & \xmark & \cmark & \xmark & \cmark & \xmark & \xmark & \cmark \\
\midrule
cutting & && && && && & \cmark & \xmark & \xmark & \cmark \\
& \citet{Lioutikov2016} & \xmark & \cmark & \xmark & \xmark & \cmark & \xmark & \cmark & \xmark \icp\icp \\
& \citet{Lenz2015b} & \cmark & \xmark & \cmark & \xmark & \xmark & \cmark & \xmark & \cmark \icp\icp \\
& \citet{Thananjeyan2017} & \xmark & \cmark & \cmark & \xmark & \cmark & \xmark & \cmark & \xmark \icp\icp \\
\midrule
peeling & \citet{Medina2017} & \xmark & \cmark & \xmark & \xmark & \xmark & \cmark & \cmark & \xmark & \cmark & \xmark & \xmark & \cmark \\
\midrule
scooping & \citet{Schenck2017} & \cmark & \xmark & \xmark & \cmark & \cmark & \xmark & \cmark & \xmark & \cmark & \xmark & \xmark & \cmark \\
\midrule
pouring &&&&&& \cmark & \xmark & \xmark & \cmark & \cmark & \xmark & \xmark & \cmark \\
& \citet{Pastor2008} & \xmark & \cmark & \cmark & \xmark \icp\icp\icp\icp \\
& \citet{Tamosiunaite2011} & \xmark & \cmark & \xmark & \xmark \icp\icp\icp\icp \\
& \citet{Brandi2014} & \cmark & \cmark & \xmark & \xmark \icp\icp\icp\icp \\
& \citet{Chi2017} & \xmark & \cmark & \xmark & \xmark \icp\icp\icp\icp \\
& \citet{Sermanet2017} & \xmark & \cmark & \xmark & \xmark \icp\icp\icp\icp \\
& \citet{Caccavale2018} & \xmark & \cmark & \cmark & \xmark \icp\icp\icp\icp \\
\midrule
collision avoidance & \citet{Koert2016} & \xmark & \cmark & \cmark & \xmark & \cmark & \xmark & \cmark & \xmark & \cmark & \xmark & \xmark & \cmark \\
\midrule
golf & \citet{Maeda2016} & \xmark & \cmark & \xmark & \xmark & \cmark & \xmark & \xmark & \cmark & \cmark & \xmark & \xmark & \cmark \\
minigolf & \citet{KhansariZadeh2012} & \xmark & \cmark & \cmark & \xmark & \cmark & \xmark & \xmark & \cmark & \cmark & \xmark & \xmark & \cmark \\
billiard & & & & & & \cmark & \xmark & \xmark & \cmark & \cmark & \xmark & \xmark & \cmark \\
& \citet{Atkeson1997} & \xmark & \cmark & \cmark & \xmark \icp\icp\icp\icp \\
& \citet{Pastor2011} & \xmark & \cmark & \xmark & \xmark \icp\icp\icp\icp \\
baseball & \citet{Peters2005} & \xmark & \cmark & \xmark & \xmark & \cmark & \xmark & \xmark & \cmark & \cmark & \xmark & \xmark & \cmark \\
badminton & \citet{Liu2013} & \xmark & \cmark & \xmark & \xmark
& \cmark & \xmark & \xmark & \cmark & \cmark & \xmark & \xmark & \cmark \\
tennis & \citet{Ijspeert2002} & \xmark & \cmark & \cmark & \xmark & \cmark & \xmark & \xmark & \cmark & \cmark & \xmark & \xmark & \cmark\\
table tennis & & & & & & \cmark & \xmark & \xmark & \cmark & \cmark & \xmark & \xmark & \cmark \\
& \citet{Kober2010} & \xmark & \cmark & \cmark & \xmark \icp\icp\icp\icp\\
& \citet{Muelling2011} & \xmark & \cmark & \cmark & \xmark \icp\icp\icp\icp\\
& \citet{Kober2012} & \xmark & \cmark & \cmark & \xmark \icp\icp\icp\icp\\
& \citet{Muelling2013} & \xmark & \cmark & \cmark & \xmark \icp\icp\icp\icp\\
tetherball & & & & & & \cmark & \xmark & \xmark & \cmark & \cmark & \xmark & \xmark & \cmark \\
& \citet{Daniel2012} & \xmark & \cmark & \xmark & \xmark
\icp\icp\icp\icp \\
& \citet{Parisi2015} & \xmark & \cmark & \cmark & \xmark \icp\icp\icp\icp \\
darts & \citet{Kober2012} & \xmark & \cmark & \xmark & \cmark & \cmark & \xmark & \xmark & \cmark & \cmark & \xmark & \xmark & \cmark \\
throwing & & & & & & \cmark & \xmark & \xmark & \cmark & \cmark & \xmark & \xmark & \cmark \\
& \citet{Gams2010} & \xmark & \cmark & \xmark & \xmark \icp\icp\icp\icp \\
& \citet{Ude2010} & \xmark & \cmark & \xmark & \xmark \icp\icp\icp\icp \\
& \citet{Kober2012} & \xmark & \cmark & \xmark & \xmark\icp\icp\icp\icp \\
& \citet{daSilva2014} & \xmark & \cmark & \xmark & \xmark \icp\icp\icp\icp \\
& \citet{Gutzeit2018} & \xmark & \cmark & \xmark & \xmark \icp\icp\icp\icp \\
kicking & & & & & & \cmark & \xmark & \xmark & \cmark & \cmark & \xmark & \xmark & \cmark \\
& \citet{Boeckmann2016} & \xmark & \cmark & \cmark & \xmark \icp\icp\icp\icp \\
& \citet{Hester2010} & \xmark & \cmark & \cmark & \xmark \icp\icp\icp\icp \\
& \citet{Asada1996} & \xmark & \cmark & \cmark & \xmark \icp\icp\icp\icp \\
\midrule
ball-in-a-cup & & & & & & \cmark & \xmark & \xmark & \cmark & \cmark & \xmark & \xmark & \cmark \\
& \citet{Kober2008} & \xmark & \cmark & \xmark & \cmark \icp\icp\icp\icp \\
& \citet{Kober2009} & \xmark & \cmark & \xmark & \cmark \icp\icp\icp\icp \\
\midrule
pancake flipping & \citet{Kormushev2010b} & \xmark & \cmark & \xmark & \cmark
& \cmark & \xmark & \xmark & \cmark & \cmark & \xmark & \xmark & \cmark \\
\midrule
nunchaku flipping & \citet{Zhao2018} & \cmark & \cmark & \xmark & \cmark & \cmark & \xmark & \xmark & \cmark & \cmark & \xmark & \xmark & \cmark \\
\midrule
archery & \citet{Kormushev2010c} & \xmark & \cmark & \xmark & \xmark & \cmark & \xmark & \cmark & \xmark & \cmark & \xmark & \xmark & \cmark \\
astrojax & \citet{Paraschos2018} & \xmark & \cmark & \xmark & \xmark & \xmark & \cmark & \xmark & \cmark & \cmark & \xmark & \xmark & \cmark \\
maracas & \citet{Paraschos2018} & \xmark & \cmark & \xmark & \xmark & \xmark & \cmark & \xmark & \cmark & \cmark & \xmark & \xmark & \cmark \\
drumming & \citet{Ude2010} & \xmark & \cmark & \xmark & \xmark & \xmark & \cmark & \xmark & \cmark & \cmark & \xmark & \xmark & \cmark \\
\midrule
balancing on wheels & \citet{Vlassis2009} & \xmark & \cmark & \cmark & \xmark & \xmark & \xmark & \xmark & \cmark & \xmark & \cmark & \xmark & \xmark \\
\midrule
postural recovery & \citet{Kuindersma2011} & \xmark & \cmark & \xmark & \xmark 
& \xmark & \xmark & \xmark & \cmark & \xmark & \cmark & \xmark & \xmark \\
\midrule
balancing inv. pendulum &&&&&& \xmark & \xmark & \xmark & \cmark & \xmark & \cmark & \xmark & \cmark \\
& \citet{Marco2016} & \xmark & \cmark & \xmark & \cmark \icp\icp\icp\icp \\
& \citet{Doerr2017} & \xmark & \cmark & \xmark & \cmark \icp\icp\icp\icp \\
\midrule
walking &&&&&& & & & & & & & \\
$\vdash$ six legs &&&&&& \xmark & \cmark & \cmark & \xmark & \cmark & \xmark & \cmark & \xmark \\
& \citet{Maes1990} & \xmark & \cmark & \xmark & \xmark \icp\icp\icp\icp  \\
& \citet{Kirchner1997} & \cmark & \cmark & \xmark & \cmark \icp\icp\icp\icp  \\
$\vdash$ quadrupedal &&&&&& \xmark & \cmark & \cmark & \xmark & \cmark & \xmark & \cmark & \xmark \\
& \citet{Birdwell2007} & \xmark & \cmark & \cmark & \xmark \icp\icp\icp\icp \\
& \citet{Kohl2004} & \xmark & \cmark & \cmark & \xmark \icp\icp\icp\icp \\
& \citet{Bartsch2016} & \xmark & \cmark & \xmark & \cmark \icp\icp\icp\icp \\
$\vdash$ biped &&&& & & \xmark & \cmark & \xmark & \cmark & \cmark & \xmark & \cmark & \xmark\\
& \citet{Benbrahim1997} & \xmark & \cmark & \xmark & \cmark \icp\icp\icp\icp \\
& \citet{Matsubara2005} & \xmark & \cmark & \xmark & \cmark \icp\icp\icp\icp \\
& \citet{Geng2006} & \cmark & \xmark & \xmark & \cmark \icp\icp\icp\icp \\
& \citet{Kormushev2011b} & \xmark & \cmark & \cmark & \xmark \icp\icp\icp\icp \\
& \citet{Missura2015} & \xmark & \cmark & \cmark & \xmark \icp\icp\icp\icp \\
\midrule
walking up stairs & \citet{Kolter2009} & \xmark & \cmark & \xmark & \xmark & \cmark & \cmark & \xmark & \cmark & \cmark & \xmark & \cmark & \xmark\\
\midrule
walking on rough terrain &&&&&&&&&& \cmark & \xmark & \cmark & \xmark \\
& \citet{Kolter2008} & \cmark & \xmark & \cmark & \xmark & \cmark & \cmark & \cmark & \xmark \icp\icp \\
& \citet{Kalakrishnan2009} & \cmark & \xmark & \cmark & \xmark & \cmark & \cmark & \cmark & \xmark \icp\icp \\
& \citet{Zucker2011} & \cmark & \xmark & \cmark & \xmark & \cmark & \cmark & \cmark & \xmark \icp\icp \\
\midrule
pedal racer & \citet{Gams2014} & \cmark & \cmark & \xmark & \cmark & \xmark & \cmark & \xmark & \cmark & \cmark & \xmark & \cmark & \xmark\\
\midrule
jumping &&&&&&&& \xmark & \cmark & \cmark & \xmark & \cmark & \xmark \\
& \citet{Kolter2009} & \xmark & \cmark & \xmark & \xmark & \cmark & \cmark \icp\icp\icp \\
& \citet{Theodorou2010} & \xmark & \cmark & \xmark & \xmark & \cmark & \xmark \icp\icp\icp \\
\midrule
dribbling & \citet{Latzke2007} & \xmark & \cmark & \cmark & \xmark & \xmark & \cmark & \xmark & \cmark & \cmark & \xmark & \cmark & \xmark\\
\midrule
standing up & \citet{Morimoto2001} & \cmark & \cmark & \xmark & \cmark & \cmark & \xmark & \xmark & \cmark & \cmark & \xmark & \cmark & \xmark \\
\midrule
collision avoidance && && && \cmark & \xmark & && \cmark & \xmark & \cmark & \xmark \\
& \citet{Tai2016b} & \cmark & \cmark & \xmark & \cmark \icp & \cmark & \xmark \icp \icp \\
& \citet{Loquercio2018} & \cmark & \xmark & \xmark & \cmark \icp & \cmark & \xmark \icp \icp \\
& \citet{Gandhi2017} & \cmark & \cmark & \xmark & \cmark \icp & \cmark & \xmark \icp \icp \\
& \citet{Kahn2017} & \cmark & \xmark & \cmark & \xmark \icp & \xmark & \cmark \icp \icp \\
\midrule
ball interception & \citet{Mueller2007} & \xmark & \cmark & \cmark & \xmark & \xmark & \xmark & \xmark & \cmark & \cmark & \xmark & \xmark & \xmark \\
defense behavior & \citet{Riedmiller2009} & \xmark & \cmark & \cmark & \xmark & \xmark & \xmark & \xmark & \cmark & \cmark & \xmark & \xmark & \xmark \\
cooperative behavior & \citet{Riedmiller2007} & \xmark & \cmark & \cmark & \xmark & \xmark & \xmark & \xmark & \cmark & \cmark & \xmark & \xmark & \xmark \\
capturing a ball & \citet{Fidelman2004} & \xmark & \cmark & \cmark & \xmark & \cmark & \xmark & \xmark & \cmark & \cmark & \xmark & \xmark & \cmark \\
\midrule
visual navigation & \citet{Zhu2017} & \cmark & \cmark & \xmark & \cmark 
& \cmark & \xmark & \cmark & \xmark & \cmark & \xmark & \cmark & \xmark \\
navigation & \citet{Silver2010} & \cmark & \xmark & \cmark & \xmark & \cmark & \xmark & \cmark & \xmark & \cmark & \xmark & \cmark & \xmark \\
navigation & \citet{Conn2007} & \cmark & \cmark & \xmark 
& \cmark 
& \cmark & \xmark & \cmark & \xmark & \cmark & \xmark & \cmark & \xmark \\
navigation & \citet{Pfeiffer2016} & \cmark & \cmark &
\cmark & 
\cmark 
& \cmark & \xmark & \cmark & \xmark & \cmark & \xmark & \cmark & \xmark \\
lane following & \citet{Chuang2018} & \cmark & \cmark & \xmark & \cmark & \cmark & \xmark & \cmark & \xmark & \cmark & \xmark & \cmark & \xmark \\
navigation and estimation & \citet{Oswald2010} & \xmark & \cmark & \cmark & \xmark & \cmark & \xmark & \xmark & \cmark & \cmark & \xmark & \cmark & \xmark \\
navigation with exploration & \citet{Cocora2006} & \xmark & \cmark & \cmark & \xmark & \cmark & \xmark & \cmark & \xmark & \cmark & \xmark & \cmark & \xmark \\
exploration & \citet{Kollar2008} & \xmark & \cmark & \cmark & \xmark & \cmark & \xmark & \cmark & \xmark & \cmark & \xmark & \cmark & \xmark \\
\midrule
active sensing & \citet{Kwok2004} & \xmark & \cmark & \cmark & \xmark & \cmark & \xmark & \cmark & \xmark & \cmark & \xmark & \xmark & \xmark \\
\midrule
unscrewing a light bulb & \citet{Manschitz2016} & \cmark & \xmark & \xmark & \cmark 
& \cmark & \xmark & \cmark & \xmark & \cmark & \xmark & \xmark & \cmark \\
\midrule
coffee / tea preparation & \citet{Caccavale2018} & \xmark & \cmark & \cmark & \xmark & \cmark & \xmark & \cmark & \cmark & \cmark & \xmark & \xmark & \cmark \\
\midrule
pizza preparation & \citet{Caccavale2017} & \xmark & \cmark & \cmark & \xmark & \cmark & \xmark & \cmark & \xmark & \cmark & \xmark & \xmark & \cmark \\
\midrule
pizza dough rolling & \citet{Figueroa2016} & \xmark & \cmark & \xmark & \cmark & \cmark & \xmark & \cmark & \xmark & \cmark & \xmark & \xmark & \cmark \\
\midrule
high five & \citet{Amor2014} & \xmark & \cmark & \xmark & \cmark & \cmark & \xmark & \xmark & \cmark & \cmark & \xmark & \xmark & \cmark\\
hand shaking & \citet{Huang2018} & \xmark & \cmark & \xmark & \xmark & \cmark & \cmark & \xmark & \cmark & \cmark & \xmark & \xmark & \cmark\\ 
\midrule
hand-over &&&&&&&&&&&&\\
& \citet{Ewerton2015} & \xmark & \cmark & \xmark & \cmark & \cmark & \xmark & \xmark & \cmark & \cmark & \xmark & \xmark & \cmark\\
& \citet{Maeda2017} & \xmark & \cmark & \xmark & \cmark & \cmark & \xmark & \xmark & \cmark & \cmark & \xmark & \xmark & \cmark\\
\midrule
holding & \citet{Ewerton2015} & \xmark & \cmark & \xmark & \cmark & \cmark & \xmark & \xmark & \cmark & \cmark & \xmark & \xmark & \cmark\\
\midrule
carrying &&&&&&&&&&&&& \\
& \citet{Rozo2015} & \cmark & \cmark & \xmark & \cmark & \cmark & \xmark & \xmark & \cmark & \xmark & \cmark & \xmark & \cmark\\
& \citet{Berger2012} & \cmark & \xmark & \xmark & \cmark & \cmark & \xmark & \xmark & \cmark & \xmark & \cmark & \xmark & \cmark\\
\midrule
lifting &&&&&&&&&&&&& \\
& \citet{Evrard2009} & \xmark & \cmark & \xmark & \cmark & \cmark & \xmark & \xmark & \cmark & \cmark & \cmark & \xmark & \cmark \\
\midrule
putting on a shoe & \citet{Canal2018} & \xmark & \cmark & \cmark & \xmark & \cmark & \xmark & \xmark & \cmark & \cmark & \xmark & \xmark & \cmark \\
\midrule
collaborative drilling & \citet{Nikolaidis2013} & \xmark & \cmark & \cmark & \xmark & \cmark & \xmark & \xmark & \cmark & \cmark & \xmark & \xmark & \cmark \\
\bottomrule
\caption*{
\dag \hspace{0.5em} \textbf{Perception and Action:} Refers to the part of the behavior that has been learned.\\
\ddag \hspace{0.5em} \textbf{Deliberative and Reactive:} Refers to the complete behavior. Behaviors are considered to be deliberative if models of the world or the robot in the world are constructed.\\
\textbf{Symbols:}\\
\begin{tabular}{ll}
$\vdash$ & Indicates that the behavior is an instance of the more general behavior
above.\\
\cmark & Behavior has this property.\\
\xmark & Behavior does not have this property.\\
& We cannot state that the behavior generally has this property.\\
\textperiodcentered & Property is inherited from the behavior category.
\end{tabular}
}
\end{longtable}
\endgroup

\pagebreak

}
%
%
{ 
%
%
%

\addcontentsline{toc}{section}{References}
\bibliographystyle{apalike}
\bibliography{literature}

} 
%
%
\newpage

\end{document}